\documentclass[10pt]{article}
\usepackage[round, authoryear]{natbib} 
\usepackage{subcaption}
\usepackage{metatemp}
\usepackage{microtype}
\usepackage{parskip}
\usepackage{titletoc}

\usepackage{booktabs}
\usepackage{caption}
\usepackage{multirow}
\usepackage{wrapfig}
\usepackage{makecell}
\usepackage{adjustbox}
\usepackage{fullpage}
\usepackage{ragged2e}

\usepackage{algorithm}
\usepackage{algorithmic}
\newtcolorbox{samplebox}[1]{
	enhanced,
	colback=LinkBlue!5!white,
	colframe=black!60,
	boxrule=0.6pt,
	arc=6pt,
	left=4pt, right=4pt, top=4pt, bottom=4pt,
	boxsep=3pt,
	before upper={\noindent\textbf{\small #1}\par\smallskip},
}

\definecolor{GenLavender}{RGB}{179, 209, 230}  
\definecolor{GenMeta}{RGB}{90,90,90}          
\definecolor{GenBad}{RGB}{214,39,40}          

\newcommand{\genhead}[2]{{\bfseries\large #1}\hfill{#2}\par\smallskip}

\newtcolorbox{genbox}[3][]{%
  enhanced, breakable,
  colback=LabPrimary!4, colframe=black!15, boxrule=0.5pt, arc=7pt,
  left=10pt, right=10pt, top=8pt, bottom=8pt,
  before skip=8pt, after skip=8pt,
  before upper={\setlength{\parskip}{0pt}\genhead{#2}{#3}},
  #1}

\newtcolorbox{genboxplain}[3][]{%
  enhanced, breakable,
  colback=white, frame hidden,
  borderline north={0.5pt}{0pt}{black!20},
  borderline south={0.5pt}{0pt}{black!20},
  sharp corners, left=2pt, right=2pt, top=8pt, bottom=8pt,
  before skip=8pt, after skip=8pt,
  before upper={\setlength{\parskip}{0pt}\genhead{#2}{#3}},
  #1}

\definecolor{GenLabel}{RGB}{0,82,150}   

\definecolor{GenLabel}{RGB}{104,82,150}   

\newlength{\genitemsep}

\newcommand{\genprompt}[1]{%
{\sffamily\bfseries\color{black}Prompt}\par\smallskip
\begingroup\setlength{\parskip}{0pt}{\color{black!80}#1}\par\endgroup
\medskip
{\color{black!15}\rule{\linewidth}{0.4pt}}\par
\medskip}

\newcommand{\genanswer}[1]{%
{\sffamily\bfseries\color{black}Answer}\par\smallskip
\begingroup\setlength{\parskip}{\genitemsep}#1\par\endgroup}

\usepackage[dvipsnames]{xcolor}

\newcommand{\arrowup}{\ensuremath{\uparrow}}

\usepackage{amsmath}
\usepackage{amssymb}
\usepackage{mathtools}
\usepackage{amsthm}
\usepackage{cancel}
\usepackage{enumitem}

\setlist[enumerate,1]{leftmargin=2em, itemsep=1em, parsep=0pt}
\setlist[itemize,1]{leftmargin=2em}

\usepackage[capitalize,noabbrev]{cleveref}
\usepackage{thm-restate}
\crefformat{section}{Section~#2#1#3}
\crefformat{subsection}{Section~#2#1#3}
\crefformat{subsubsection}{Section~#2#1#3}
\crefformat{equation}{(#2#1#3)}
\crefrangeformat{equation}{(#3#1#4) to (#5#2#6)}
\crefmultiformat{equation}{(#2#1#3)}{ and (#2#1#3)}{, (#2#1#3)}{ and (#2#1#3)}
\crefname{appendix}{Appendix}{Appendices}
\Crefname{appendix}{Appendix}{Appendices}
\crefformat{appendix}{Appendix~#2#1#3}
\AddToHook{cmd/appendix/before}{%
\crefalias{section}{appendix}%
\crefalias{subsection}{appendix}%
\crefalias{subsubsection}{appendix}
}
\crefname{table}{Table}{Tables}
\crefname{figure}{Figure}{Figures}
\crefname{lemma}{Lemma}{Lemmas}
\crefname{assumption}{Assumption}{Assumptions}

\definecolor{d62728}{RGB}{214,39,40}
\definecolor{1f77b4}{RGB}{31,119,180}
\definecolor{9467bd}{RGB}{148,103,189}
\definecolor{FigNY}{HTML}{1a1a3e}    
\definecolor{FigSD}{HTML}{b0a8d0}    
\usepackage{colortbl}
\definecolor{pinegreen}{rgb}{0.0, 0.47, 0.44}
\definecolor{cornellred}{rgb}{0.7, 0.11, 0.11}
\definecolor{cadmiumgreen}{rgb}{0.0, 0.42, 0.24}
\definecolor{spirodiscoball}{rgb}{0.06, 0.75, 0.99}
\definecolor{blizzardblue}{rgb}{0.73, 0.96, 0.99}
\definecolor{aliceblue}{rgb}{0.91, 0.94, 0.97}
\definecolor{darkblue}{RGB}{232, 224, 240} 
\definecolor{Red7}{rgb}{0.941, 0.243, 0.243}
\definecolor{Green7}{RGB}{55, 178, 77}

\makeatletter
\def\thm@space@setup{%
  \thm@preskip=0.8em plus 0.2em minus 0.2em
  \thm@postskip=0.8em plus 0.2em minus 0.2em
}
\makeatother

\theoremstyle{plain}
\newtheorem{theorem}{Theorem}[section]
\newtheorem{proposition}[theorem]{Proposition}

\theoremstyle{definition}

\tcolorboxenvironment{proposition}{
  metatemp-common,
  colback=LabPrimary!4,
  left=12pt,
  right=12pt,
  top=8pt,
  bottom=8pt
}

\crefname{proposition}{Proposition}{Propositions}
\Crefname{proposition}{Proposition}{Propositions}

\usepackage{pifont}
\usepackage{array}
\usepackage{xurl}

\newcommand{\R}{\mathbb{R}}

\newcolumntype{C}[1]{>{\centering\arraybackslash}p{#1}}
\newcolumntype{L}[1]{>{\raggedright\arraybackslash}p{#1}}

\usepackage[textsize=tiny]{todonotes}

\title{Masked Language Flow Models}

\author{Iskander Azangulov}
\author[\star]{Kianoosh Ashouritaklimi}
\author[\star]{Leo Zhang}
\author{Simon Vary}
\author{Patrick Rebeschini}
\affiliation{Department of Statistics, University of Oxford}
\affiliation[\star]{Equal contribution}

\begin{document}
\begin{abstract}
\looseness=-1
\emph{Masked Diffusion Models} (MDMs) promise fast, parallel language generation, but their reverse transition factorises across token positions---an approximation that breaks down in the few-step sampling regime where parallel generation ought to provide the greatest efficiency gains.
\emph{Flow Language Models} (FLMs) sidestep this limitation by learning a continuous flow that transports noise toward clean sequences represented in Euclidean space, inducing a flow map that can be distilled for single-step generation. 
However, this makes complex tasks requiring multi-step reasoning problematic for FLMs, as FLMs are forced to decode every token during generation.
To address this, we introduce \textbf{Masked Language Flow Models (MLFMs)}, which incorporate masking into FLMs using a continuous stochastic interpolant to bridge partially masked and clean sequences.
This design enables conditional generation via continuous flows and allows pretrained MDMs to be converted into MLFMs through a simple, lightweight adaptation.
Leveraging this flexibility, we propose a novel sampler that alternates continuous denoising with the discrete unmasking of confident tokens to better support multi-step reasoning.
We evaluate our approach on GSM8K and MT-Bench and find, for the first time, that flow-based language models can be scaled to solve downstream reasoning and instruction-following tasks.

\vspace{0.5cm}
    Code is available at: \href{https://github.com/imbirik/mlfm}{github.com/imbirik/mlfm}.

\end{abstract}
\maketitle

\section{Introduction}

\emph{Autoregressive Models} (ARMs) have driven much of the recent progress in language modelling by framing language generation as next-token prediction~\citep{brown2020languagemodelsfewshotlearners}. 
However, their left-to-right factorisation of the joint distribution makes this process inherently sequential, as each token must be conditioned on all preceding ones. 
Consequently, inference costs scale linearly with output length, creating a bottleneck for long sequences. 
This limitation has motivated growing interest in \emph{Masked Diffusion Models} (MDMs) ~\citep{austin2021structured, campbell2022continuous, lou2024discreteratios,shi2024simplified, sahoo2024simple} which replace sequential generation with the parallel decoding of masked tokens.

\looseness=-1
Although MDMs have demonstrated strong performance across language modelling and downstream tasks~\citep{nie2025scalingmdm, nie2025lldm}, they typically rely on a factorised reverse transition across masked positions, making inference tractable in the combinatorially large discrete state space. 
This approximation is accurate in the infinitesimal-step limit but becomes highly restrictive in the few-step regime~\citep{deschenaux2024beyond}, where each step must independently decode many masked tokens at once---ignoring the dependencies amongst them. 
Consequently, the very regimes where parallel decoding promises the greatest speedups~\citep{dieleman2023language, zheng2024masked} are precisely those where this independence assumption most severely compromises generation quality.

\emph{Flow Language Models} (FLMs) \citep[][]{roos2026categoricalflowmaps,lee2026flowmaplanguagemodels,potaptchik2026discreteflowmaps,davis2026scalingcategoricalflowmaps,chen2026langflow, hu2026elfembeddedlanguageflows} address this issue by moving from a discrete state space into a continuous one.
Specifically, FLMs learn a flow \citep{song2020score, lipman2022flow, albergo2023stochastic} transporting noise to continuous embeddings of token sequences, allowing latent states to evolve jointly across token positions.
A significant advantage of FLMs is that they also induce a flow map which can be distilled to support few-step and even one-step generation.
However, collapsing generation to a single flow map may be too restrictive for language tasks that require multi-step reasoning.
Indeed, many language tasks benefit from iterative generation where partially completed intermediate states provide context for subsequent refinements~\citep{ghazvininejad-etal-2019-mask,nye2021show,wei2022chain}.

To address this limitation, we propose \textbf{Masked Language Flow Models (MLFMs)}. MLFMs incorporate masking from MDMs into FLMs by using a Brownian bridge as a stochastic interpolant connecting embedded partially masked sequences with embedded clean sequences.  As a result, MLFMs retain the exact any-position conditioning structure of MDMs while learning a coupled continuous flow for generating clean tokens at masked positions. 

Moreover, this special structure of MLFMs gives us a natural strategy for training and inference. For training, we propose adapting pretrained MDMs into MLFMs: this is possible because, at the boundary of the Brownian bridge, the stochastic interpolant reduces to an embedded partially masked sequence, matching the setup of MDMs. This allows us to warm-start MLFM training from large pretrained MDMs, greatly reducing the compute required to train our models.

\looseness=-1
For inference, we leverage the flexibility MLFMs provide through any-position conditional generation. Specifically, we introduce a novel sampling scheme that combines a new approach to classifier-free guidance with the online promotion of confident tokens. Under this guidance method, clean tokens are noised in the reference velocity to isolate their contribution. Concurrently, if the posterior mode at a specific token position reaches a probability of at least $1-\epsilon$, we immediately commit to that token and substitute in its clean embedding. By promoting these resolved tokens early, they instantly become useful context for subsequent generation steps, rather than remaining noisy until the final step. 

Empirically, we adapt a pretrained MDM with 1028M parameters from \cite{nie2025scalingmdm} into an MLFM and evaluate the resulting model on GSM8K \citep{cobbe2021gsm8k} and MT-Bench \citep{zheng2023judging}. MLFM improves over SMDM and similarly sized AR baselines on MT-Bench and obtains encouraging performance on GSM8K, demonstrating its ability to handle both instruction-following and reasoning tasks. Across these benchmarks, we also find that our novel sampling scheme provides significant improvements. To the best of our knowledge, this is the first time that flow-based language models have been scaled to downstream reasoning and instruction-following tasks.

\section{Background}

In this section, we provide an overview of the background required for our approach. Section \ref{sub:mdm} discusses Masked Diffusion Models and Section \ref{sub:flm} discusses Flow Language Models, in particular LangFlow \citep{chen2026langflow} which we take as the basis for our approach.

\paragraph{Notation.}

Throughout, we let \(\mathcal V\) denote a vocabulary of tokens, \(|\mathcal V|\) the vocabulary size, and \(L\) the sequence length.
We will identify elements in $\mathcal{V}$ with the natural numbers: $\{1, \ldots, |\mathcal{V}|\}$.
A (clean) sequence $X\sim p$, where $p$ denotes our data distribution, is written as
\(X=(X^1,\ldots,X^L)\in\mathcal V^L\).  
We use \(s\in[0,1]\) for the masking probability and \(t\in[0,1]\) for the continuous noising time.

\subsection{Masked Diffusion Models}
\label{sub:mdm}
\emph{Masked Diffusion Models} (MDMs) construct a non-autoregressive generative model for $p$ by introducing a special \emph{mask} token \texttt{[MASK]} to the vocabulary $\mathcal{V}$ and defining a continuous-time Markov chain (CTMC) \citep{del2017stochastic} transporting $p$ to the Dirac measure on fully masked sequences.
The forward process $X_s \sim p_{s|0}(X_s| X)$ of the CTMC at the masking probability $s\in[0, 1]$ and $X\sim p$ is given by the following sampling procedure:
\begin{equation}    
 \label{eq:mdm_forward}
    X_s^\ell  =
    \begin{cases}
        X^\ell, & \text{ if }B^\ell=0,\\
        \texttt{[MASK]}, & \text{ if } B^\ell=1,
    \end{cases}
    \qquad
    \text{ where } B^\ell\stackrel{\mathrm{i.i.d.}}{\sim}\operatorname{Bernoulli}(s).
\end{equation}
We let $\mathcal M_s = \{\ell: X_s^\ell =  \texttt{[MASK]}\}$ denote the positions of mask tokens in $X_s$ and $\mathcal U_s = \{\ell: X_s^\ell \ne\texttt{[MASK]}\}$ the positions of clean tokens in $X_s$.

We can generate approximate samples from $p$ by simulating the backward process of the CTMC. This involves gradually unmasking tokens to reveal clean tokens which are then fixed for later steps. This relies on access to the ground-truth factorised posterior $p_{\text{data}}^\ell(X^\ell|X_s^{\mathcal U_s})$ of clean tokens given unmasked tokens \citep{zheng2024masked, ou2025your} where $X_s^{\mathcal{U}_s}$ denotes the subcollection of $X_s$ consisting of clean tokens.
This can be learned from data via the following likelihood-based objective: 
\begin{equation}
\label{eq:mdm_loss}
    \mathcal L_{\mathrm{MDM}}(\theta)
    =
    \mathbb E_{X\sim p, X_s \sim p_{s|0}(\cdot|X), s\sim U(0, 1)}
    \left[
        \frac{1}{s}
        \sum_{\ell\in \mathcal M_s}
        -\log p_\theta^\ell(X^\ell\mid X_s^{\mathcal U_s})
    \right],
\end{equation}
where we parametrise $p_\theta^\ell$ with a neural network to approximate $p_\text{data}^\ell$ and $U(0, 1)$ denotes the uniform distribution over $[0,1]$.

During inference, MDMs can unmask multiple tokens in parallel in a single step.  This is most effective when the tokens to be unmasked are close to conditionally independent given the current masked state, due to the factorised structure of $p_\theta^\ell$.
However, this is only typically true for a small subset of tokens as increasing the number of tokens to be unmasked increases the likelihood of strong dependencies between tokens. 
As a result, MDMs still require many refinement steps for high quality samples despite their parallel decoding structure.

\subsection{Flow Language Models}
\label{sub:flm}

We consider the \emph{LangFlow}~\citep{chen2026langflow} approach to \emph{Flow Language Models} (FLMs) which operates within a learnt embedding space instead of one--hot embeddings.

Let \(H\) be the embedding dimension.
For each token $a\in\mathcal{V}$ in the vocabulary, we represent the token with its corresponding embedding vector \(E_a \in \R^H\) and we let \(E\in\mathbb R^{|\mathcal V|\times H}\) denote the overall embedding matrix.
Moreover, for a sequence $X\sim p$, we represent $X$ in Euclidean space by the embedding:
\[
    z_1=(E_{X^1},\ldots,E_{X^L})\in\mathbb R^{L\times H}.
\]
\looseness=-1
With this continuous representation $z_1$ of discrete sequences, we can construct a generative model for $p$ by considering the Gaussian--based stochastic interpolant:
\begin{equation}
\label{eq:flm_interpolant}
    z_t = t z_1 + (1-t)\epsilon,
\end{equation}
where $\epsilon\sim\mathcal N(0, I_{L\times H})$ and $t\in[0, 1]$. 
We note that $z_0$ is Gaussian noise and $z_1$ is an embedded sequence from $p$.
To sample from the flow induced by this stochastic interpolant, we parametrise the denoiser $p_\phi$ which approximates the factorised posterior over clean tokens:
\[
    p_\phi(\cdot\mid z_t,t)
    \in\Delta(\mathcal V)^L
    \subset\mathbb R^{L\times|\mathcal V|}.
\]
This is trained with the following cross-entropy objective:
\begin{equation}
\label{eq:flm_ce}
    \mathcal L_{\mathrm{FLM}}(\phi)
    =
    \mathbb E_{t\sim\pi_t, X, \epsilon}
    \left[
        \frac{1}{L}
        \sum_{\ell=1}^L
        -\log p_\phi^\ell(X^\ell\mid z_t,t)
    \right],
\end{equation}
where $\pi_t$ denotes some distribution over $[0, 1]$. Although the supervision target \(X^\ell\) is discrete, the learned
denoiser induces a continuous endpoint estimate:
\begin{equation}
\label{eq:embedding_posterior_mean}
    \widehat z_{1,\phi}(z_t,t)
    :=
    P_\phi(z_t,t)E,
    \qquad \text{ where } \qquad
    P_\phi(z_t,t)=p_\phi(\cdot\mid z_t,t)
    \in\mathbb R^{L\times|\mathcal V|},
\end{equation}
and \(\widehat z_{1,\phi}(z_t,t)\) estimates the posterior
mean embedding \(\mathbb E[z_1\mid z_t,t]\).
Additionally, \(\widehat z_{1,\phi}(z_t,t)\) allows us to estimate the velocity field for the flow:
\begin{equation}
\label{eq:flm_velocity}
    v_\phi(z_t,t)
    :=
    \frac{\widehat z_{1,\phi}(z_t,t)-z_t}{1-t},
    \qquad t<1.
\end{equation}
Therefore, we can generate approximate samples from $p$ by integrating the ODE:
\[
    \frac{d}{dt} z_t=v_\phi(z_t,t).
\]
In practice, the ODE is discretized over a finite time grid, and the
final token sequence is obtained by decoding the denoiser probabilities from $p_\phi$ in
vocabulary space.

Continuous flows are attractive because they define a joint trajectory from
noise to clean sequences. 
This trajectory can be distilled into a direct flow map or a consistency-style solver \citep{song2023consistency, boffi2026build}, reducing the number of generation steps.
However, an exact one-shot map is an overly demanding object for language models---i.e. some problems (such as maths or coding) may naturally require iterative reasoning in which intermediate commitments are used as context for later decisions.

\section{Masked Language Flow Models}

We introduce the core framework of MLFMs in Section~\ref{sub:framework}. 
In Section \ref{sub:mdm_to_mlfm}, we show how pretrained MDMs can naturally be adapted to MLFMs, including the architectural changes required for this adaptation.
Finally, Section~\ref{sub:conditional_fine_tuning} describes how we can apply supervised fine-tuning on instruction-response data to MLFMs, an aspect that, to the best of our knowledge, has not been explored in prior work on FLMs. 

\subsection{Framework} \label{sub:framework}

We retain the use of mask tokens and token embeddings $E$ from MDMs and FLMs.
We begin by defining a forward process $z_{s,t} \in \mathbb{R}^{L \times H}$ in Euclidean space. 
This process is parametrised by $(s,t) \in [0,1]^2$ where $s$ denotes the masking probability and $t$ denotes the noising time.

Let \(m=E_{\mathtt{[MASK]}}\) denote the embedding of the $\texttt{[MASK]}$ token.  For $X\sim p$, $s \sim \pi_s$, $t \sim \pi_t$, where $\pi_s$, $\pi_t$ are distributions on $[0,1]$, we sample a partially masked sequence $X_s\sim p_{s|0}(\cdot|X)$ as in the MDM forward process \eqref{eq:mdm_forward} and we construct $z_{s,t} \in \R^{L\times H}$ as follows. For all positions \(\ell\in\mathcal U_s\) corresponding to clean tokens in $X_s$, we fix $z^\ell_{s,t}$ to be the clean embedding:
\[
    z_{s,t}^\ell=E_{X^\ell},
    \qquad \forall \ell\in\mathcal U_s, \  t\in[0,1].
\]
For $\ell\in\mathcal M_s$, we construct $z_{s,t}^\ell$ via the stochastic interpolant formed by the Brownian bridge connecting $m$ and $E_{X^\ell}$:
\begin{equation}
\label{eq:mlfm_forward}
\begin{aligned}
    z_{s,t}^\ell\mid X^\ell
    \sim
    \mathcal N
    \Big(
        (1-t)m+tE_{X^\ell},\quad\!\!\!\!
        \sigma^2t(1-t)I_H
    \Big),
    \qquad \qquad \forall \ell\in\mathcal M_s, \  t\in[0,1],
\end{aligned}
\end{equation}
\looseness=-1
where $\sigma>0$ is some choice of noise scale on masked positions. 
Since the variance vanishes at both endpoints, we see that the stochastic interpolant satisfies $z_{s,0}^\ell=m$ and $z_{s,1}^\ell=E_{X^\ell}$ almost surely.
Thus $z_{s,0}$ recovers the embedding of the partially masked sequence $X_s$ and $z_{s,1}$ recovers the embedding of $X$.
Additionally, for $t\in(0,1)$, the masked positions are modelled as noisy continuous states that provide partial information about their underlying clean token embeddings.

\looseness=-1
This construction is related to recent hybrid discrete-continuous  diffusion models for language \citep{pynadath2025candi, zheng2025continuously, zhou2026coevolutionarycontinuousdiscretediffusion} which also combine discrete masking and continuous noising. The aforementioned works construct a forward process that masks and noises tokens simultaneously with the motivation that continuous noising provides additional information to help scaffold the generation of discrete tokens.  
However, in MLFMs, we \emph{decouple} these components---i.e. for any choice of sequence masking, we learn a continuous flow to generate clean tokens to fill in the masked tokens.
This separation is distillation-friendly and enables the flexible inference procedures described in Section~\ref{sec:sampling}, where tokens can be revealed adaptively rather than at a fixed rate, as well as our adaptation of pretrained MDMs to accelerate training. 

Following the same reasoning as in \cite{chen2026langflow}, we sample from the flow induced by this stochastic interpolant by parametrising the MLFM denoiser $p_\theta$ to output token probabilities from the factorised posterior over clean tokens:
\[
    p_\theta(\cdot\mid z_{s,t},t)
    \in\Delta(\mathcal V)^L
    \subset\mathbb R^{L\times|\mathcal V|}.
\]
 This is trained via cross-entropy on only \emph{masked} subsets of tokens:
\begin{equation}
\label{eq:mlfm_loss}
    \mathcal L_{\mathrm{MLFM}}(\theta)
    =
    \mathbb E_{s,t,X, \epsilon}
    \left[
        \frac{1}{|\mathcal M_s|}
        \sum_{\ell\in\mathcal M_s}
        -\log p_\theta^\ell(X^\ell\mid z_{s,t},t)
    \right].
\end{equation}
As in \cite{chen2026langflow}, we find that the choice of $\pi_t$ is critical for performance. We therefore follow their entropy-based schedule, with a small adaptation to our setting. Further details are provided in Appendix~\ref{app:gamma_fitting}, in the context of the experimental setup in Section~\ref{sec:experiments}.

For sampling, we have that $p_\theta$ induces an estimate of the posterior mean embedding
\begin{equation}
\label{eq:mlfm_endpoint_estimate}
    \widehat z_{1,\theta}(z_{s,t},t)
    :=
    P_\theta(z_{s,t},t)E,
    \qquad
    P_\theta(z_{s,t},t)
    =
    p^\ell_\theta(\cdot\mid z_{s,t},t),
\end{equation}
which can then be used to estimate the velocity field of the flow:
\begin{equation}
\label{eq:mlfm_velocity}
    v_\theta^\ell(z_{s,t},t)
    :=
    \begin{cases}
        \displaystyle
        \frac{
            \widehat z_{1,\theta}^\ell(z_{s,t},t)-z_{s,t}^\ell
        }{1-t},
        & \ell\in\mathcal M_s,\\[1.5ex]
        0,
        & \ell\in\mathcal U_s,
    \end{cases}
    \qquad t<1.
\end{equation}
\looseness=-1
With this, we can generate samples by integrating the ODE defined by $v_\theta$. This evolves only the masked positions, while the positions in $\mathcal U_s$ remain clamped to their clean embeddings. Note that the incorporation of masks naturally enables \emph{conditional} generation in MLFMs, which has received limited attention in prior work on FLMs. Moreover, this additional flexibility allows us to design more advanced sampling schemes for MLFMs, as discussed in Section~\ref{sec:sampling}.

\subsection{MDMs to MLFMs} \label{sub:mdm_to_mlfm}

It is easy to see that we can view the MLFM loss as a continuous extension of the MDM loss: at the endpoint $t=0$, every unresolved position $\ell \in M_s$ is represented exactly by the mask embedding $m$, while every observed position $\ell \in U_s$ is kept at its clean embedding. Thus, $z_{s,0}$ contains the same information as the partially masked sequence $X_s$. 
Consequently, the cross-entropy objective $\mathcal{L}_{\mathrm{MLFM}}$ at $t=0$ has the same minimiser as the MDM objective $\mathcal{L}_{\mathrm{MDM}}$: in both cases, the optimal predictor is the factorised posterior distribution of each masked token given the clean tokens.
We formalise this endpoint equivalence in the following proposition.
\begin{proposition}[Mask-endpoint equivalence]
\label{prop:mask_endpoint_equivalence}
Assume that the unmasked token identities are either provided directly or are
recoverable from their embeddings under \(E\).  At the mask endpoint \(t=0\),
the MLFM prediction problem coincides with the MDM prediction problem: for any
\(\ell\in\mathcal M_s\) and \(a\in\mathcal V\),
\begin{equation}
\label{eq:mask_endpoint_equivalence}
    \mathbb P(X^\ell=a\mid z_{s,0})
    =
    \mathbb P(X^\ell=a\mid X_s^{\mathcal U_s},\mathcal M_s)
    =
    \mathbb P(X^\ell=a\mid X_s).
\end{equation}
\begin{proof}
See Appendix \ref{proof:mask_endpoint_equivalence}.
\end{proof}

\end{proposition}

\looseness=-1
This observation motivates adapting pretrained MDMs to MLFMs: because a pretrained MDM has already learned the denoising problem at the endpoint \(z_{s,0}\), it provides a strong initialisation for MLFM training and avoids the cost of training a capable MLFM from scratch.
Concretely, we adapt a pretrained bidirectional-transformer from an MDM for MLFM training by
adapting three main components in the architecture:
\begin{enumerate}[label=\roman*)]
    \item The MDM token embedding layer \(E\in\mathbb R^{|\mathcal V|\times H}\), which maps tokens into the embedding space, is used as the MLFM embedding matrix and is kept fixed during adaptation, similar to~\cite{hu2026elfembeddedlanguageflows}. 
    \item The MDM transformer blocks initialise the MLFM denoising
    transformer and are frozen.  We augment the normalisations in these blocks with AdaLN \citep{nie2025scalingmdm} to condition on the continuous corruption time \(t\), and attach LoRA adapters \citep{hu2022lora} to the linear layers to adapt the pretrained transformer to work with continuous corrupted embeddings. 
    \item The MDM head which maps the final activations to  vocabulary logits is adapted with its own LoRA adapter.
\end{enumerate}

\looseness=-1
With these architectural changes, we can continue training the pretrained MDM under the MLFM objective $\mathcal{L}_\mathrm{MLFM}(\theta)$, thereby adapting it into an MLFM.
We refer to this MDM-to-MLFM training procedure as \emph{adaptation} and we provide a summary in Algorithm \ref{alg:mlfm_pretraining_step}.

\subsection{Supervised Fine-Tuning}
\label{sub:conditional_fine_tuning}

In this section, we describe how to perform supervised fine-tuning (SFT) with MLFMs.
Let $\mathcal D_{\mathrm{FT}}=\{(p_i,a_i)\}_{i=1}^n$ denote a SFT dataset, where \(p_i\) is a prompt and \(a_i\) is the corresponding target answer.
After tokenisation, let \(X=X(p_i,a_i)\) denote the sequence obtained from the concatenation of the prompt and answer in a suitable format and let \(\mathcal P\) denote the positions of the prompt and \(\mathcal A\) the positions of the answer in $X$.
We fine-tune our MLFM $p_\theta$ with the below cross-entropy objective:
\begin{equation}
\label{eq:mflm_finetuning_loss}
    \mathcal L_{\mathrm{FT}}(\theta)
    =
    \mathbb E_{(p,a)\sim\mathcal D_{\mathrm{FT}}, s\sim\pi_s, t\sim\pi_t, z_{s,t}}
    \left[
        \frac{1}{|\mathcal M_s|}
        \sum_{\ell\in\mathcal M_s}
        -\log p_\theta^\ell(X^\ell\mid z_{s,t},t)
    \right]. 
\end{equation}
For a sequence $X$ constructed from $(p,a)$, we form $z_{s,t}$ as follows. With probability $\alpha$, we mask all answer positions in $\mathcal A$, and with probability $1-\alpha$ we instead sample a subset of answer positions by applying $p_{s|0}$ only to $\mathcal A$. The prompt positions $\mathcal P$ are always kept clean to encourage conditional generation from the prompt. After the masked positions $\mathcal M_s$ are sampled, we apply the same noising process \eqref{eq:mlfm_forward} at time $t$ to these positions.
We provide a summary of our SFT procedure in Algorithm \ref{alg:mlfm_sft_step}.

\section{Sampling from MLFMs}
\label{sec:sampling}
In this section, we design a sampling scheme that exploits the additional flexibility of MLFMs for conditioning and guidance. The main idea behind our sampler is that high-confidence token predictions are more useful as clean context than as noisy latent states. In a flow-only sampler, each generated position remains latent until the final decoding step, even when its posterior distribution has already concentrated on a single token. MLFMs can use such positions more effectively: once a token becomes highly probable, we \emph{promote} it to the observed-token state, so that later denoising steps condition on its clean embedding rather than on a corrupted representation of the same position.

We can naturally combine this strategy with a variant of classifier-free guidance \citep{ho2022classifier}. Specifically, we compare two predicted vector fields: a guided field, which conditions on the promoted tokens as clean observed context, and a reference field, which uses the same tokens and positions but corrupts their embeddings to the current time. The difference between these fields isolates the effect of clean context on the model's predicted direction. We then sample from the guided flow while promoting high-confidence tokens online, yielding a single sampling procedure that combines FLM sampling with MDM-style unmasking.

We begin by recalling the DDPM sampler in Section \ref{sec:ddpm_sampler}, then introduce our variant of classifier-free guidance (Section \ref{sec:cfg}) and  online promotion strategy (Section \ref{sub:online_token_promotion}).

\subsection{DDPM Sampler}
\label{sec:ddpm_sampler}

We first describe the standard DDPM sampler \citep{ho2020denoising} for conditional generation. Let
\(\mathcal U_0\subseteq[L]\) be the unmasked context positions and let
\(\mathcal M_0=[L]\setminus\mathcal U_0\) be the masked positions we want to generate.
The sampler is initialised at the masked endpoint $z_0^\ell$:
\[
    z_0^\ell
    =
    \begin{cases}
        E_{X^\ell}, & \ell\in\mathcal U_0,\\
        m, & \ell\in\mathcal M_0 .
    \end{cases}
\]
During sampling, the context positions remain fixed, so that \(z_t^\ell=E_{X^\ell}\) for
\(\ell\in\mathcal U_0\), while the masked positions in $\mathcal M_0$ are jointly denoised by simulating the SDE
\[
d z^\ell_{t}
    =
    v^\ell_\theta(z_{s,t},t)\,dt
    +
    \sigma\,dW_t^\ell, \quad \ell\in \mathcal{M}_0.
\]
We discretize this process over a mesh \(0=t_0<t_1<\cdots<t_N=1\). Between  \(t_i\) and \(t_{i+1}\), the process is approximated using the DDPM transition 
\begin{equation}
\label{eq:ddpm_step}
    z_{t_{i+1}}^\ell\mid z_{t_i}
    \sim
    \mathcal N
    \left(
        \frac{1-t_{i+1}}{1-t_i}z_{t_i}^\ell
        +
        \frac{t_{i+1}-t_i}{1-t_i}
        \widehat z_{1}^\ell(z_{t_i},t_i),
        \,
        \sigma^2
        \frac{(t_{i+1}-t_i)(1-t_{i+1})}{1-t_i}
        I_H
    \right),
\end{equation}
which samples from the Brownian bridge connecting $z_{t_i}$ to the current endpoint estimate $\widehat z_{1}(z_{t_i},t_i)$, given by
\[
    \widehat z_{1}(z_{t_i},t_i)
    =
    z_{t_i}
    +
    (1-t_i)v_\theta(z_{t_i},t_i).
\]
Finally, at \(t_N=1\), each generated position \(\ell\) is decoded by sampling
\(X^\ell \sim p_\theta^\ell(\cdot\mid z_{t_N},t_N)\), yielding the sample \(X\).

\subsection{Classifier-Free Guidance With Corrupted Context}
\label{sec:cfg}

We first describe our variant of classifier-free guidance mentioned earlier. Let \(z_t\) be the state at time \(t\).  For $\ell\in\mathcal U_0$, the observed token is represented, as before, by its clean embedding,$$z_t^\ell=E_{X^\ell}, \qquad \ell\in\mathcal U_0.$$
We construct a corrupted-context reference state $z_{t,\mathrm{corr}}$ by applying the forward corruption marginal in~\eqref{eq:mlfm_forward} only to context positions, while leaving all other positions unchanged:
$$z_{t,\mathrm{corr}}^\ell = \begin{cases} \widetilde z_t^\ell, & \ell\in\mathcal U_0,\\ z_t^\ell, & \ell\notin\mathcal U_0 , \end{cases}$$
where, for $\ell\in\mathcal U_0$,$$\widetilde z_t^\ell\mid X^\ell \sim \mathcal N \Big( (1-t)m+tE_{X^\ell},\quad \sigma^2t(1-t)I_H \Big).$$

Let $v_\theta$ denote the MLFM velocity defined in~\eqref{eq:mlfm_velocity}. We define context-corrupted guidance with scale $w$ by
$$v_w(z_t,t) = v_\theta(z_t,t) + w\bigl( v_\theta(z_t,t) - v_\theta(z_{t,\mathrm{corr}},t) \bigr).$$
The difference term captures the effect of clean context on the predicted velocity: both model calls use the same tokens at the same positions, but only $v_\theta(z_t,t)$ observes the context through exact clean embeddings. In effect, this encourages the sampler to follow directions that are specific to the clean-context prediction, rather than directions that persist when the context is corrupted. Note that both $z_t$ and $z_{t,\mathrm{corr}}$ remain on the support of the forward process. We refer to this sampling strategy as \emph{context-corrupted classifier-free guidance} (CCFG) and summarise it in Algorithm~\ref{alg:corrupted_context_cfg}.

\begin{algorithm}[t]
\caption{Context Corrupted Classifier-Free Guidance  (CCFG)}
\label{alg:corrupted_context_cfg}
\begin{algorithmic}[1]
\REQUIRE State \(z_t\), time \(t\), clean token positions
\(\mathcal U_0\), clean tokens \(X^{\mathcal U_0}\), guidance scale \(w\).
\STATE Initialize \(z_{t,\mathrm{corr}}\leftarrow z_t\).
\FOR{\(\ell\in\mathcal U_0\)}
    \STATE Draw \(\widetilde z_t^\ell\mid X^\ell\) from the corruption
    marginal in~\eqref{eq:mlfm_forward}.
    \STATE Set \(z_{t,\mathrm{corr}}^\ell\leftarrow\widetilde z_t^\ell\).
\ENDFOR
\STATE \textbf{return}
\(v_w=v_\theta(z_t,t)+
w\bigl(v_\theta(z_t,t)-v_\theta(z_{t,\mathrm{corr}},t)\bigr)\).
\end{algorithmic}
\end{algorithm}

\subsection{Online Token Promotion}
\label{sub:online_token_promotion}
\begin{algorithm}[!t]
\caption{CCFG with Online Token Promotion (CCFG w/ OTP)}
\label{alg:flow_guided_unmasking}
\begin{algorithmic}[1]
\REQUIRE Prompt tokens \(X^{\mathcal U_0}\), unresolved positions
\(\mathcal M_0\), mesh \(0=t_0<\cdots<t_N=1\), guidance scale \(w\),
tolerance \(\varepsilon\).
\STATE Initialize \(z_{t_0}\) with clean 
embeddings on \(\mathcal U_0\) and mask embeddings on \(\mathcal M_0\).
\FOR{\(i=0,\ldots,N-1\)}
    \STATE Set
    \(\widehat X_i^\ell=\operatorname*{arg\,max}_{a\in\mathcal V}
    p_\theta^\ell(a\mid z_{t_i},t_i)\) for \(\ell\in\mathcal M_i\).
    \STATE Promote
    \(\mathcal P_i=\{\ell\in\mathcal M_i:
    p_\theta^\ell(\widehat X_i^\ell\mid z_{t_i},t_i)\ge 1-\epsilon\}\).
        \STATE For \(\ell\in \mathcal P_i\), set \(X^\ell=\widehat X_i^\ell\) and
    \(z_{t_i}^\ell=E_{X^\ell}\).
    \STATE Set \(\mathcal U_{i+1}=\mathcal U_i\cup \mathcal P_i\) and
    \(\mathcal M_{i+1}=\mathcal M_i\setminus P_i\).
    \STATE Compute guided velocity $v_w$ using~\cref{alg:corrupted_context_cfg}.
    \STATE Make one DDPM step~\eqref{eq:ddpm_step} on \(\mathcal M_{i+1}\) using
    \(v_w\) to obtain \(z_{t_{i+1}}\).
\ENDFOR
\STATE Decode any remaining unresolved positions and \textbf{return} \(X\).
\end{algorithmic}
\end{algorithm}

The standard DDPM sampler, as well as CCFG, treat the prompt as clean context but keep every generated position latent until the final decoding step. This can be inefficient
as different positions resolve at different times: indeed, positions adjacent to
the prompt, deterministic formatting tokens, and padding tokens in short
answers often have sharply peaked posteriors well before \(t=1\). If such positions remain latent, however, later model calls still observe them through corrupted representations rather than through the clean embeddings of the predicted tokens.

We therefore \emph{promote} high-confidence positions online, fixing a token to its
clean embedding as soon as the model is sufficiently confident in it. Formally, let
$\mathcal{U}_i$ and $\mathcal{M}_i=[L]\setminus\mathcal{U}_i$ denote the set of unmasked, context
and masked positions at step $i$, respectively. For each masked position, the posterior
\(p_\theta(\cdot\mid z_{t_i},t_i)\) gives the predicted token
\[
    \widehat X_i^\ell
    =
    \operatorname*{arg\,max}_{a\in\mathcal V}
    p_\theta^\ell(a\mid z_{t_i},t_i),
\]
and, given a tolerance \(\varepsilon>0\), we promote the positions
\[
    \mathcal P_i
    =
    \left\{
        \ell\in\mathcal M_i:
        p_\theta^\ell(\widehat X_i^\ell\mid z_{t_i},t_i)
        \ge 1-\varepsilon
    \right\}.
\]

For each \(\ell\in\mathcal P_i\) we set \(X^\ell=\widehat X_i^\ell\) and fix
\(z_{t_i}^\ell=E_{X^\ell}\). We then apply the DDPM transition~\eqref{eq:ddpm_step} only to the still-masked positions, holding all
positions in \(\mathcal U_i\cup\mathcal P_i\) fixed, and update the sets $\mathcal U_i$ and   $\mathcal M_i$:
\[
    \mathcal U_{i+1}=\mathcal U_i\cup\mathcal P_i,
    \qquad
    \mathcal M_{i+1}=\mathcal M_i\setminus\mathcal P_i .
\]
The sampler terminates once $t_N=1$ is reached or all positions have been
promoted. We call this sampling strategy \emph{online token promotion} (OTP) and in practice combine it with CCFG.  Algorithm~\ref{alg:flow_guided_unmasking} summarises our full sampling procedure.
Additionally, we note that this sampling procedure can be further accelerated with the use of flow map distillation as promotion can be done at any time $t$.


\subsubsection{Error from Online Token Promotion}
\label{sec:err_online_token_promotion}
We note that OTP can introduce errors into the sampling process by promoting tokens too early. Indeed, even when the posterior mode has high probability, it may still be incorrect, causing the sampler to fix the position to the clean embedding of an incorrect token before terminal time. The following result upper bounds the error caused by such early promotions.

\begin{proposition}[Promotion Error]
\label{prop:promotion_error}
Let \(p\) be the target distribution on \(\mathcal V^L\), and assume that each
denoising samples exactly from \(p(z_{t_{i+1}}\mid z_{t_i})\).
Let \(\widetilde p\) be the output law of the corresponding sampler that uses
the promotion rule above with the true posteriors under \(p\), and then
continues with the same exact denoising dynamics conditioned on promoted
values.  Then
\[
    \operatorname{TV}(p,\widetilde p)\le \varepsilon L .
\]

\begin{proof}
See Appendix \ref{proof:promotion_error}.
\end{proof}
\end{proposition}

The above results show that the overall accumulated probability of at least one incorrect promotion is bounded by \(\varepsilon L\). Note that this error does not depend on the number of discretization steps, and can be made arbitrarily small by taking $\epsilon$ small enough.

As a result, we can view Online Token Promotion as a signal amplifier, similar to self-conditioning~\citep{chen2023analogbitsgeneratingdiscrete}. We note that this is not available to previous work in hybrid discrete-continuous approaches to language modelling~\citep{pynadath2025candi,zheng2025continuously, zhou2026coevolutionarycontinuousdiscretediffusion}.

\section{Experiments}
\label{sec:experiments}

Most prior work on FLMs has evaluated unconditional generation, typically using metrics such as generative perplexity and entropy. While these metrics aim to measure distributional modelling quality, they are unreliable indicators of actual performance \citep{franca2026hacking} and they do not establish whether flow-based language models can serve as useful conditional generators in practical settings. We therefore evaluate MLFM on more demanding downstream tasks that require mathematical reasoning and instruction following. Concretely, we adapt the pretrained SMDM model of \cite{nie2025scalingmdm} into an MLFM, following Section \ref{sub:framework}, and evaluate it on GSM8K \citep{cobbe2021gsm8k} and MT-Bench \citep{zheng2023judging}.

Below, we first describe our experimental setup in Section \ref{sec:general_experimental_setup} then present our main results and ablations in Sections \ref{sec:main_results} and \ref{sec:ablations} respectively.

\subsection{Experimental Setup}
\label{sec:general_experimental_setup}
\subsubsection{Adaptation Setup}
\label{sec:experimental_setup}

\looseness=-1
For all of our experiments, we initialise our model from a pretrained SMDM with 1028M parameters\footnote{More specifically, we use the official SMDM 1028M checkpoint,
\texttt{mdm-1028M-3300e18-rsl-0.01-bs-1024.safetensors}, from the
\url{https://huggingface.co/nieshen/SMDM} Hugging Face repository.}. The pretrained backbone and input embedding matrix are kept frozen throughout all training stages. We train only a small set of adapters: LoRA adapters on attention and MLP modules, an output-head LoRA adapter, and AdaLN adapters for time conditioning. The backbone LoRA rank is 256 with $\alpha=512$ and dropout 0.05; the output-head LoRA rank is 256 with $\alpha=256$ and no dropout. The MLP that produces the AdaLN time-conditioning parameters has hidden dimension 512. The total number of trainable parameters in our model is 319M. For further details, see Appendix \ref{app:additional_exp_detials}.

\looseness=-1
We note that aside from \cite{davis2026scalingcategoricalflowmaps}, who train a 1.7B-parameter categorical flow model at trillion-token scale, MLFM is, to the best of our knowledge, the first flow-language model at billion-parameter scale.

\subsubsection{Adaptation Training}
\label{sec:adaptation_training}

We train our model on SlimPajama \citep{cerebras2023slimpajama}, tokenized with the LLaMA-2 tokenizer \citep{touvron2023llama} using sequences of length 1024. We use a training budget of $\approx$100B processed sequence positions---this corresponds to 200k optimiser updates with an effective batch size of 512 sequences, obtained by accumulating two global batches of 256 sequences. We use the AdamW optimiser \citep{loshchilov2017decoupled} with 3k warmup steps followed by cosine decay to 10\% of the peak learning rate. The learning rate is $10^{-4}$ for the backbone LoRA, output-head LoRA, and AdaLN parameters. We apply weight decay of 0.01 and clip the global gradient norm at 1.0. We also maintain an exponential moving average (EMA) of the adapter weights with decay 0.999, and use these EMA weights for validation.

Moreover, following \cite{chen2026langflow}, we sample the time $t$ in terms of the log noise-to-signal ratio (NSR) $\gamma$ rather than sampling $t$ directly. We sample $\gamma$ from a three-component mixture consisting of uniform, normal, and fitted generalised-logistic components, with mixture weights $0.1/0.2/0.7$ respectively (see Appendix \ref{app:gamma_fitting} for further details). For our distribution $\pi_s$ over masking probabilities $s$, we use the MaskGIT cosine schedule \citep{chang2022maskgit} clipped to $[0.05,1.0]$. For our bridge \eqref{eq:mlfm_forward}, we set $\sigma=0.2$. 
In addition, we use an auxiliary embedding loss with weight $10$: from the predicted token distribution, we form the corresponding posterior-mean embedding and penalise its distance to the clean token embedding. Our adaptation stage took approximately three days on 16 NVIDIA GH200 GPUs.

\subsubsection{Supervised Fine-Tuning}
\label{sec:sft}

After the adaptation stage, we further train the adapted MLFM model on supervised fine-tuning data consisting of prompt-response pairs. The data mixture includes general instruction data from first-turn ShareGPT\footnote{The dataset can be accessed from \url{https://sharegpt.com/}.} examples, mathematical reasoning data from NuminaMath-CoT \citep{numina_math_datasets}, GSM8K-Aug-NL \citep{cobbe2021gsm8k,deng2023implicit}, and MetaMathQA \citep{yu2024metamath}, and code data from OpenCodeInstruct \citep{ahmad2025opencodeinstruct}. We sample these three groups with weights 0.2, 0.6, and 0.2, respectively.

As discussed in Section \ref{sub:conditional_fine_tuning}, prompt tokens are always kept fixed during supervised fine-tuning and the MLFM objective is applied only to response tokens. With probability 0.5, all response tokens are masked; otherwise, the response mask ratio is sampled using the same MaskGIT cosine schedule as in Section \ref{sec:adaptation_training}. We use an SFT budget of $\approx$15B processed sequence positions. We use AdamW for 50k optimizer updates, using an effective batch size of 512 sequences obtained by accumulating two global batches of size 256. We use 3k warmup steps followed by cosine learning-rate decay, with peak learning rate $5\times10^{-5}$ for the LoRA, output-head LoRA, and AdaLN parameters. Our SFT took  $\approx$18 hours on 16 NVIDIA GH200 GPUs. 

\subsubsection{Datasets}
\label{sec:datasets}
We evaluate our SFT model on two conditional generation benchmarks: GSM8K \citep{cobbe2021gsm8k} and MT-Bench \citep{zheng2023judging}. GSM8K is a benchmark of grade-school mathematical word problems, where each example consists of a natural-language question with a numerical answer. We use GSM8K to evaluate mathematical reasoning, and measure performance using exact-match accuracy after extracting the final numerical answer from the model output. 

MT-Bench is an open-ended instruction-following benchmark consisting of multi-turn user queries spanning diverse categories \citep{zheng2023judging}. Following standard practice, model responses are instead evaluated by a strong language-model judge, which assigns a scalar score reflecting the quality of the answer. In our experiments, we report the first-turn MT-Bench score with GPT-4o \citep{achiam2023gpt} as our judge.

For both datasets, we prompt the model using the Vicuna prompt template~\citep{vicuna2023}.

\subsubsection{Baselines}
\label{sec:baselines}
We compare our approach with the 1.1B MDM model of \cite{nie2025scalingmdm}, which we refer to as SMDM, as well as the AR models considered in their work: LLaMA-2 \citep{touvron2023llama} for GSM8K, and their similarly sized AR model for MT-Bench.

For inference, we follow Algorithm \ref{alg:flow_guided_unmasking} with $\epsilon=0.05$. We use a maximum sequence length of 512 tokens and 256 sampling steps for GSM8K, and 1024 tokens and 128 sampling steps for MT-Bench. Similar to \cite{nie2025scalingmdm}, we report the results with the best guidance scale $w \in \{0, 2, 4, 6\}$ and ablate with different $w$ in Section \ref{sec:ablations}.

\subsection{Main Results}
\label{sec:main_results}
Table~\ref{tab:main_results} shows that MLFM improves substantially over both baselines on MT-Bench, achieving a first-turn score of $2.27$ compared with $1.60$ for SMDM and $1.57$ for the similarly sized AR baseline. Notably, this gain is obtained with $128$ sampling steps, half of the $256$ steps used by SMDM. On GSM8K, however, MLFM remains significantly behind both LLaMA-2 and SMDM, obtaining $31.24\%$ accuracy compared with $58.6\%$ and $58.5\%$, respectively. One plausible reason is the difference in fine-tuning protocol: the SMDM GSM8K result is obtained after task-specific fine-tuning on augmented GSM8K data for $40$ epochs, whereas our MLFM is fine-tuned on a broader instruction mixture. Nevertheless, these results are encouraging: to the best of our knowledge, they provide the first evidence that flow-based language models can be scaled beyond unconditional generation to downstream reasoning and instruction-following tasks. The qualitative examples in Figures \ref{fig:gen-example-mt-bench}, \ref{fig:gen-example-gsm8k},  provide further evidence that MLFM can produce coherent, multi-step responses. 

\begin{table}[!h]
\centering
\caption{Main results on GSM8K and MT-Bench comparing MLFM with SMDM and the AR and LLaMA-2 baselines of \cite{nie2025scalingmdm}. Note that SMDM uses 256 sampling steps for both datasets.}
\label{tab:main_results}
\begin{tabular}{lcc}
\toprule
Approach 
& \makecell{GSM8K\\(accuracy\% $\arrowup$)}
& \makecell{MT-Bench\\(first-turn score $\arrowup$)} \\
\midrule
LLaMA-2 \citep{touvron2023llama} & 58.6 & -- \\
AR baseline \citep{nie2025scalingmdm}                 & -- & 1.57 \\
SMDM \citep{nie2025scalingmdm}                 & 58.5 & 1.60 \\
\hline
\hline
MLFM                 & 31.24 & 2.27 \\
\bottomrule
\end{tabular}
\end{table}

\subsection{Ablations}
\label{sec:ablations}
\begin{figure}[!h]
    \centering
    \makebox[0.5\linewidth]{\textbf{MT--Bench}}
    \vspace{0.05em}
    
    \begin{subfigure}[b]{0.3\linewidth}
        \centering
        \includegraphics[width=\linewidth]{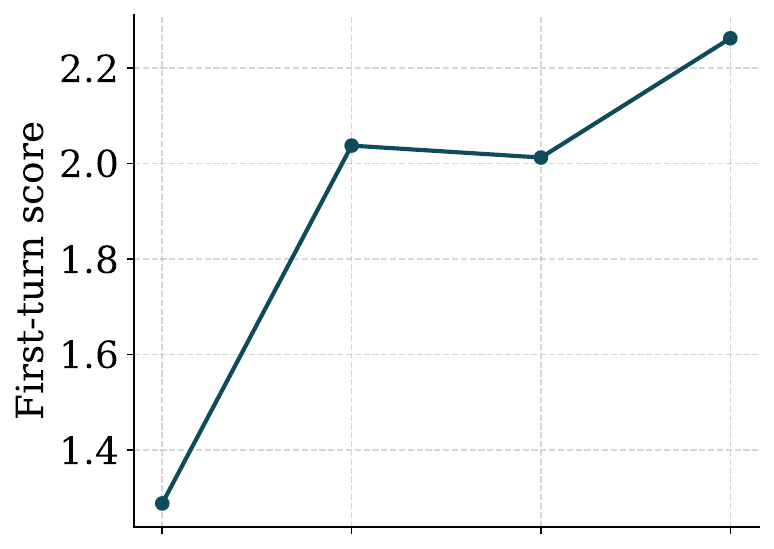}
    \end{subfigure}    
    \hspace{1em}
    \begin{subfigure}[b]{0.3\linewidth}
        \centering
        \includegraphics[width=\linewidth]{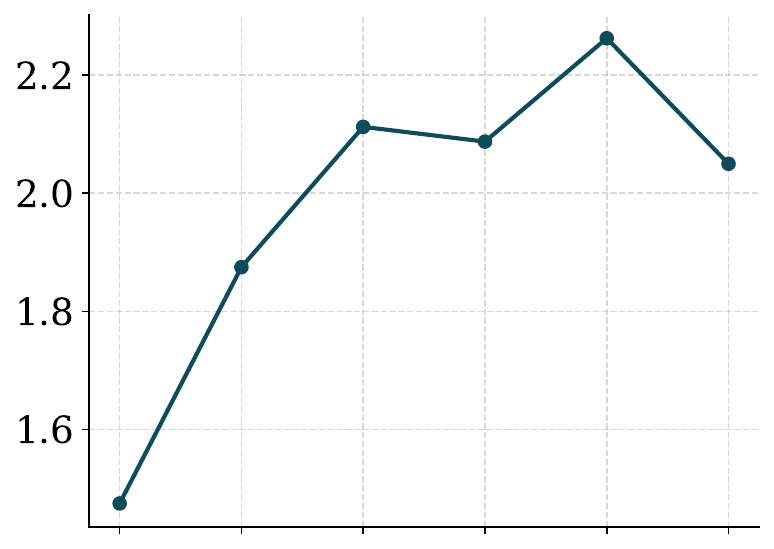}
    \end{subfigure}
    
    \vspace{1em}     
    \makebox[0.5\linewidth]{\textbf{GSM8K}}
    \vspace{0.05em}
    
    \begin{subfigure}[b]{0.3\linewidth}
        \centering
        \includegraphics[width=\linewidth]{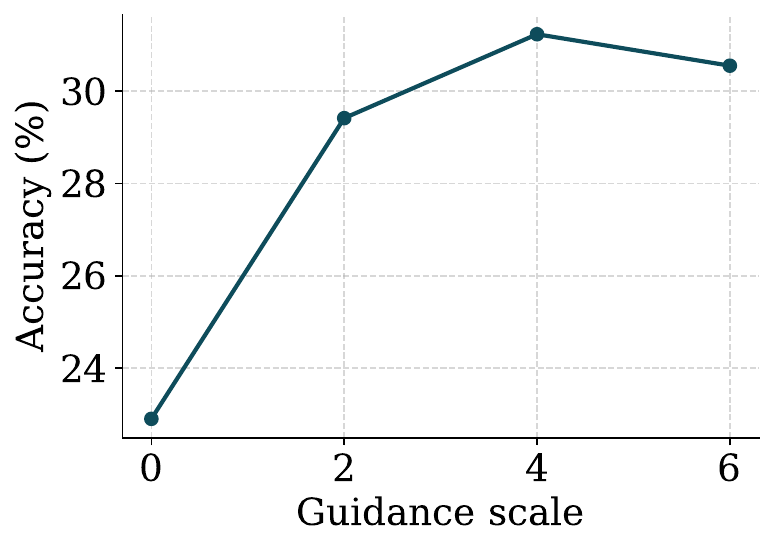} 
    \end{subfigure}    
    \hspace{1em}
    \begin{subfigure}[b]{0.3\linewidth}
        \centering
        \includegraphics[width=\linewidth]{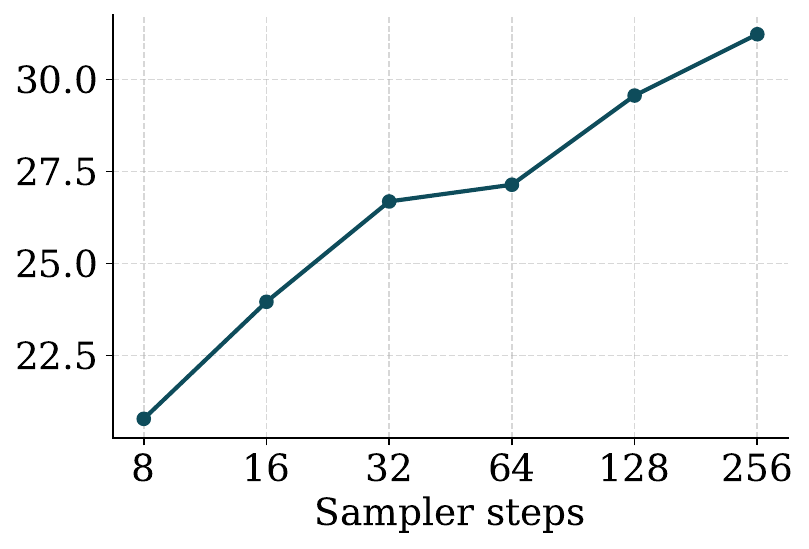} 
    \end{subfigure}    
    \caption{MT-Bench and GSM8K results across different guidance scales (left) and sampler steps (right). The plots on the right use the optimal guidance scales given by the corresponding plots on the left.}
    \label{fig:ablations}
\end{figure}

Here, we study the effect of different guidance scales $w$, different numbers of sampling steps, and different sampling strategies. 

\paragraph{Different sampling strategies.}
Table \ref{tab:ablation_sampling} shows the GSM8K and MT-Bench results across three sampling settings: no guidance (the standard DDPM sampler from Section \ref{sec:ddpm_sampler}), CCFG (Algorithm \ref{alg:corrupted_context_cfg}), and CCFG with Online Token Promotion (Algorithm \ref{alg:flow_guided_unmasking}). On both datasets, we observe that both CCFG and online token promotion significantly boost performance. This is not surprising: OTP commits high-confidence posterior modes as clean observed tokens, giving later denoising steps more reliable context than corrupted continuous states. 

\begin{table}[!h]
\centering
\caption{Results for GSM8K and MT-Bench for different sampling strategies.}
\label{tab:ablation_sampling}
\begin{tabular}{lcc}
\toprule
Sampling 
& \makecell{GSM8K\\(accuracy\% $\arrowup$)}
& \makecell{MT-Bench\\(first-turn score $\arrowup$)} \\
\midrule
No guidance  & 13.19 & 1.22 \\
CCFG (Algorithm \ref{alg:corrupted_context_cfg})                  & 21.38 & 1.85 \\
\makecell[l]{CCFG w/ OTP (Algorithm \ref{alg:flow_guided_unmasking})}                  & 31.24 & 2.27 \\
\bottomrule
\end{tabular}
\end{table}

\paragraph{Different guidance scales.} Figure \ref{fig:ablations} shows the GSM8K and MT-Bench results for different guidance scales. In general, we see that larger guidance scales provide the largest gains in performance. This too is not surprising, as stronger guidance makes the sampler rely more heavily on the clean observed context when resolving the remaining tokens.

\paragraph{Different numbers of sampling steps.} Similarly, Figure \ref{fig:ablations} shows the GSM8K and MT-Bench results for different numbers of sampling steps. We see that, in general, larger numbers of sampling steps provide the largest gains in performance. Moreover, we note that MLFM still outperforms the SMDM and AR baselines on MT-Bench even at 16 sampling steps.

\section{Conclusion}
\label{sec:conclusion}

\looseness=-1
In this work, we introduced Masked Language Flow Models which integrate masking from Masked Diffusion Models into Flow Language Models via a Brownian bridge connecting partially masked sequences with clean sequences. This enables exact, any-position conditional generation, allowing MLFMs to anchor continuous generation on partially masked sequences. Paired with our novel sampler, this facilitates complex, multi-step reasoning. Additionally, MLFMs support efficient training with a lightweight adaptation of pretrained MDMs. For future work, it is interesting to continue scaling MLFMs as well as distilling such models.

\newpage

\section*{Acknowledgments}

\looseness=-1
IA, KA and LZ would like to thank Jinwoo Kim and Pete Patterson for helpful conversations.

IA is supported by the Engineering and Physical Sciences Research Council [grant number EP/T517811/1]. 
LZ and KA are supported by the EPSRC CDT in Modern Statistics and Statistical Machine Learning (EP/S023151/1).
This work was supported by the UKRI AI Research Resource (AIRR) through Isambard-AI (project AIRR-GW - Diffusion Models for Language Modelling) and by an Amazon Research Award awarded to Patrick Rebeschini (Fall 2024). SV and PR are funded by UK Research and Innovation (UKRI) under the UK government’s Horizon Europe funding guarantee [grant number EP/Y028333/1].

\bibliographystyle{plainnat}
\bibliography{main}

\begin{thebibliography}{49}
\providecommand{\natexlab}[1]{#1}
\providecommand{\url}[1]{\texttt{#1}}
\expandafter\ifx\csname urlstyle\endcsname\relax
  \providecommand{\doi}[1]{doi: #1}\else
  \providecommand{\doi}{doi: \begingroup \urlstyle{rm}\Url}\fi

\bibitem[Achiam et~al.(2023)Achiam, Adler, Agarwal, Ahmad, Akkaya, Aleman, Almeida, Altenschmidt, Altman, Anadkat, et~al.]{achiam2023gpt}
Josh Achiam, Steven Adler, Sandhini Agarwal, Lama Ahmad, Ilge Akkaya, Florencia~Leoni Aleman, Diogo Almeida, Janko Altenschmidt, Sam Altman, Shyamal Anadkat, et~al.
\newblock Gpt-4 technical report.
\newblock \emph{arXiv preprint arXiv:2303.08774}, 2023.

\bibitem[Ahmad et~al.(2025)Ahmad, Ficek, Samadi, Huang, Noroozi, Majumdar, and Ginsburg]{ahmad2025opencodeinstruct}
Wasi~Uddin Ahmad, Aleksander Ficek, Mehrzad Samadi, Jocelyn Huang, Vahid Noroozi, Somshubra Majumdar, and Boris Ginsburg.
\newblock Opencodeinstruct: A large-scale instruction tuning dataset for code llms.
\newblock \emph{arXiv preprint arXiv:2504.04030}, 2025.

\bibitem[Albergo et~al.(2025)Albergo, Boffi, and Vanden-Eijnden]{albergo2023stochastic}
Michael Albergo, Nicholas~M Boffi, and Eric Vanden-Eijnden.
\newblock Stochastic interpolants: A unifying framework for flows and diffusions.
\newblock \emph{Journal of Machine Learning Research}, 26\penalty0 (209):\penalty0 1--80, 2025.

\bibitem[Austin et~al.(2021)Austin, Johnson, Ho, Tarlow, and Van Den~Berg]{austin2021structured}
Jacob Austin, Daniel~D Johnson, Jonathan Ho, Daniel Tarlow, and Rianne Van Den~Berg.
\newblock Structured denoising diffusion models in discrete state-spaces.
\newblock \emph{Advances in neural information processing systems}, 34:\penalty0 17981--17993, 2021.

\bibitem[Boffi et~al.(2026)Boffi, Albergo, and Vanden-Eijnden]{boffi2026build}
Nicholas Boffi, Michael Albergo, and Eric Vanden-Eijnden.
\newblock How to build a consistency model: Learning flow maps via self-distillation.
\newblock \emph{Advances in Neural Information Processing Systems}, 38:\penalty0 33346--33382, 2026.

\bibitem[Brown et~al.(2020)Brown, Mann, Ryder, Subbiah, Kaplan, Dhariwal, Neelakantan, Shyam, Sastry, Askell, Agarwal, Herbert-Voss, Krueger, Henighan, Child, Ramesh, Ziegler, Wu, Winter, Hesse, Chen, Sigler, Litwin, Gray, Chess, Clark, Berner, McCandlish, Radford, Sutskever, and Amodei]{brown2020languagemodelsfewshotlearners}
Tom~B. Brown, Benjamin Mann, Nick Ryder, Melanie Subbiah, Jared Kaplan, Prafulla Dhariwal, Arvind Neelakantan, Pranav Shyam, Girish Sastry, Amanda Askell, Sandhini Agarwal, Ariel Herbert-Voss, Gretchen Krueger, Tom Henighan, Rewon Child, Aditya Ramesh, Daniel~M. Ziegler, Jeffrey Wu, Clemens Winter, Christopher Hesse, Mark Chen, Eric Sigler, Mateusz Litwin, Scott Gray, Benjamin Chess, Jack Clark, Christopher Berner, Sam McCandlish, Alec Radford, Ilya Sutskever, and Dario Amodei.
\newblock Language models are few-shot learners, 2020.
\newblock URL \url{https://arxiv.org/abs/2005.14165}.

\bibitem[Campbell et~al.(2022)Campbell, Benton, De~Bortoli, Rainforth, Deligiannidis, and Doucet]{campbell2022continuous}
Andrew Campbell, Joe Benton, Valentin De~Bortoli, Thomas Rainforth, George Deligiannidis, and Arnaud Doucet.
\newblock A continuous time framework for discrete denoising models.
\newblock \emph{Advances in Neural Information Processing Systems}, 35:\penalty0 28266--28279, 2022.

\bibitem[Chang et~al.(2022)Chang, Zhang, Jiang, Liu, and Freeman]{chang2022maskgit}
Huiwen Chang, Han Zhang, Lu~Jiang, Ce~Liu, and William~T Freeman.
\newblock Maskgit: Masked generative image transformer.
\newblock In \emph{Proceedings of the IEEE/CVF conference on computer vision and pattern recognition}, pages 11315--11325, 2022.

\bibitem[Chen et~al.(2023)Chen, Zhang, and Hinton]{chen2023analogbitsgeneratingdiscrete}
Ting Chen, Ruixiang Zhang, and Geoffrey Hinton.
\newblock Analog bits: Generating discrete data using diffusion models with self-conditioning, 2023.
\newblock URL \url{https://arxiv.org/abs/2208.04202}.

\bibitem[Chen et~al.(2026)Chen, Liang, Sui, Guo, Cheng, You, and Liu]{chen2026langflow}
Yuxin Chen, Chumeng Liang, Hangke Sui, Ruihan Guo, Chaoran Cheng, Jiaxuan You, and Ge~Liu.
\newblock Langflow: Continuous diffusion rivals discrete in language modeling.
\newblock \emph{arXiv preprint arXiv:2604.11748}, 2026.

\bibitem[Chiang et~al.(2023)Chiang, Li, Lin, Sheng, Wu, Zhang, Zheng, Zhuang, Zhuang, Gonzalez, Stoica, and Xing]{vicuna2023}
Wei-Lin Chiang, Zhuohan Li, Zi~Lin, Ying Sheng, Zhanghao Wu, Hao Zhang, Lianmin Zheng, Siyuan Zhuang, Yonghao Zhuang, Joseph~E. Gonzalez, Ion Stoica, and Eric~P. Xing.
\newblock Vicuna: An open-source chatbot impressing gpt-4 with 90\%* chatgpt quality, March 2023.
\newblock URL \url{https://lmsys.org/blog/2023-03-30-vicuna/}.

\bibitem[Cobbe et~al.(2021)Cobbe, Kosaraju, Bavarian, Chen, Jun, Kaiser, Plappert, Tworek, Hilton, Nakano, Hesse, and Schulman]{cobbe2021gsm8k}
Karl Cobbe, Vineet Kosaraju, Mohammad Bavarian, Mark Chen, Heewoo Jun, Lukasz Kaiser, Matthias Plappert, Jerry Tworek, Jacob Hilton, Reiichiro Nakano, Christopher Hesse, and John Schulman.
\newblock Training verifiers to solve math word problems.
\newblock \emph{arXiv preprint arXiv:2110.14168}, 2021.

\bibitem[Davis et~al.(2026)Davis, Filippova, Ablin, Turrisi, Shidani, Cuturi, and Béthune]{davis2026scalingcategoricalflowmaps}
Oscar Davis, Anastasiia Filippova, Pierre Ablin, Victor Turrisi, Amitis Shidani, Marco Cuturi, and Louis Béthune.
\newblock Scaling categorical flow maps, 2026.
\newblock URL \url{https://arxiv.org/abs/2605.07820}.

\bibitem[Del~Moral and Penev(2017)]{del2017stochastic}
Pierre Del~Moral and Spiridon Penev.
\newblock \emph{Stochastic processes: From applications to theory}.
\newblock Chapman and Hall/CRC, 2017.

\bibitem[Deng et~al.(2023)Deng, Prasad, Fernandez, Smolensky, Chaudhary, and Shieber]{deng2023implicit}
Yuntian Deng, Kiran Prasad, Roland Fernandez, Paul Smolensky, Vishrav Chaudhary, and Stuart Shieber.
\newblock Implicit chain of thought reasoning via knowledge distillation.
\newblock \emph{arXiv preprint arXiv:2311.01460}, 2023.

\bibitem[Deschenaux and Gulcehre(2024)]{deschenaux2024beyond}
Justin Deschenaux and Caglar Gulcehre.
\newblock Beyond autoregression: Fast llms via self-distillation through time.
\newblock \emph{arXiv preprint arXiv:2410.21035}, 2024.

\bibitem[Dieleman(2023)]{dieleman2023language}
Sander Dieleman.
\newblock Diffusion language models.
\newblock \url{https://benanne.github.io/2023/01/09/diffusion-language.html}, 2023.
\newblock Accessed: 2026-01-25.

\bibitem[Franca and Tong(2026)]{franca2026hacking}
Antonio Franca and Alexander Tong.
\newblock Hacking generative perplexity: Why unconditional text evaluation needs distributional metrics.
\newblock \emph{arXiv preprint arXiv:2606.08417}, 2026.

\bibitem[Ghazvininejad et~al.(2019)Ghazvininejad, Levy, Liu, and Zettlemoyer]{ghazvininejad-etal-2019-mask}
Marjan Ghazvininejad, Omer Levy, Yinhan Liu, and Luke Zettlemoyer.
\newblock Mask-predict: Parallel decoding of conditional masked language models.
\newblock In Kentaro Inui, Jing Jiang, Vincent Ng, and Xiaojun Wan, editors, \emph{Proceedings of the 2019 Conference on Empirical Methods in Natural Language Processing and the 9th International Joint Conference on Natural Language Processing (EMNLP-IJCNLP)}, pages 6112--6121, Hong Kong, China, November 2019. Association for Computational Linguistics.
\newblock \doi{10.18653/v1/D19-1633}.
\newblock URL \url{https://aclanthology.org/D19-1633/}.

\bibitem[Ho and Salimans(2022)]{ho2022classifier}
Jonathan Ho and Tim Salimans.
\newblock Classifier-free diffusion guidance.
\newblock \emph{arXiv preprint arXiv:2207.12598}, 2022.

\bibitem[Ho et~al.(2020)Ho, Jain, and Abbeel]{ho2020denoising}
Jonathan Ho, Ajay Jain, and Pieter Abbeel.
\newblock Denoising diffusion probabilistic models.
\newblock \emph{Advances in neural information processing systems}, 33:\penalty0 6840--6851, 2020.

\bibitem[Hu et~al.(2022)Hu, Shen, Wallis, Allen-Zhu, Li, Wang, Wang, Chen, et~al.]{hu2022lora}
Edward~J Hu, Yelong Shen, Phillip Wallis, Zeyuan Allen-Zhu, Yuanzhi Li, Shean Wang, Liang Wang, Weizhu Chen, et~al.
\newblock Lora: Low-rank adaptation of large language models.
\newblock \emph{Iclr}, 1\penalty0 (2):\penalty0 3, 2022.

\bibitem[Hu et~al.(2026)Hu, Qiu, Lu, Zhao, Li, Kim, Andreas, and He]{hu2026elfembeddedlanguageflows}
Keya Hu, Linlu Qiu, Yiyang Lu, Hanhong Zhao, Tianhong Li, Yoon Kim, Jacob Andreas, and Kaiming He.
\newblock Elf: Embedded language flows, 2026.
\newblock URL \url{https://arxiv.org/abs/2605.10938}.

\bibitem[Kingma et~al.(2021)Kingma, Salimans, Poole, and Ho]{kingma2021variational}
Diederik Kingma, Tim Salimans, Ben Poole, and Jonathan Ho.
\newblock Variational diffusion models.
\newblock \emph{Advances in neural information processing systems}, 34:\penalty0 21696--21707, 2021.

\bibitem[Lee et~al.(2026)Lee, Yoo, Agarwal, Shah, Huang, Raghunathan, Hong, Boffi, and Kim]{lee2026flowmaplanguagemodels}
Chanhyuk Lee, Jaehoon Yoo, Manan Agarwal, Sheel Shah, Jerry Huang, Aditi Raghunathan, Seunghoon Hong, Nicholas~M. Boffi, and Jinwoo Kim.
\newblock Flow map language models: One-step language modeling via continuous denoising, 2026.
\newblock URL \url{https://arxiv.org/abs/2602.16813}.

\bibitem[LI et~al.(2024)LI, Beeching, Tunstall, Lipkin, Soletskyi, Huang, Rasul, Yu, Jiang, Shen, Qin, Dong, Zhou, Fleureau, Lample, and Polu]{numina_math_datasets}
Jia LI, Edward Beeching, Lewis Tunstall, Ben Lipkin, Roman Soletskyi, Shengyi~Costa Huang, Kashif Rasul, Longhui Yu, Albert Jiang, Ziju Shen, Zihan Qin, Bin Dong, Li~Zhou, Yann Fleureau, Guillaume Lample, and Stanislas Polu.
\newblock Numinamath.
\newblock \url{[https://huggingface.co/AI-MO/NuminaMath-CoT](https://github.com/project-numina/aimo-progress-prize/blob/main/report/numina_dataset.pdf)}, 2024.

\bibitem[Lipman et~al.(2023)Lipman, Chen, Ben-Hamu, Nickel, and Le]{lipman2022flow}
Yaron Lipman, Ricky T.~Q. Chen, Heli Ben-Hamu, Maximilian Nickel, and Matthew Le.
\newblock Flow matching for generative modeling.
\newblock In \emph{The Eleventh International Conference on Learning Representations}, 2023.
\newblock URL \url{https://openreview.net/forum?id=PqvMRDCJT9t}.

\bibitem[Loshchilov and Hutter(2017)]{loshchilov2017decoupled}
Ilya Loshchilov and Frank Hutter.
\newblock Decoupled weight decay regularization.
\newblock \emph{arXiv preprint arXiv:1711.05101}, 2017.

\bibitem[Lou et~al.(2024)Lou, Meng, and Ermon]{lou2024discreteratios}
Alex Lou, Chenlin Meng, and Stefano Ermon.
\newblock Discrete diffusion modeling by estimating the ratios of the data distribution.
\newblock \emph{arXiv preprint arXiv:2310.16834}, 2024.

\bibitem[Nie et~al.(2025{\natexlab{a}})Nie, Zhu, Du, Pang, Liu, Zeng, Lin, and Li]{nie2025scalingmdm}
Shengqi Nie, Fenglin Zhu, Chengpeng Du, Tianyu Pang, Qi~Liu, Gang Zeng, Min Lin, and Chenguang Li.
\newblock Scaling up masked diffusion models on text.
\newblock \emph{arXiv preprint arXiv:2410.18514}, 2025{\natexlab{a}}.

\bibitem[Nie et~al.(2025{\natexlab{b}})Nie, Zhu, You, Zhang, Ou, Hu, Zhou, Lin, Wen, and Li]{nie2025lldm}
Shengqi Nie, Fenglin Zhu, Zhen You, Xin Zhang, Jing Ou, Jing Hu, Jun Zhou, Yichang Lin, Ji-Rong Wen, and Chenguang Li.
\newblock Large language diffusion models.
\newblock \emph{arXiv preprint arXiv:2502.09992}, 2025{\natexlab{b}}.

\bibitem[Nye et~al.(2021)Nye, Andreassen, Gur-Ari, Michalewski, Austin, Bieber, Dohan, Lewkowycz, Bosma, Luan, et~al.]{nye2021show}
Maxwell Nye, Anders~Johan Andreassen, Guy Gur-Ari, Henryk Michalewski, Jacob Austin, David Bieber, David Dohan, Aitor Lewkowycz, Maarten Bosma, David Luan, et~al.
\newblock Show your work: Scratchpads for intermediate computation with language models.
\newblock 2021.

\bibitem[Ou et~al.(2025)Ou, Nie, Xue, Zhu, Sun, Li, and Li]{ou2025your}
Jingyang Ou, Shen Nie, Kaiwen Xue, Fengqi Zhu, Jiacheng Sun, Zhenguo Li, and Chongxuan Li.
\newblock Your absorbing discrete diffusion secretly models the conditional distributions of clean data.
\newblock In \emph{International Conference on Learning Representations}, volume 2025, pages 64972--65009, 2025.

\bibitem[Peebles and Xie(2023)]{peebles2023scalable}
William Peebles and Saining Xie.
\newblock Scalable diffusion models with transformers.
\newblock In \emph{Proceedings of the IEEE/CVF international conference on computer vision}, pages 4195--4205, 2023.

\bibitem[Potaptchik et~al.(2026)Potaptchik, Yim, Saravanan, Holderrieth, Vanden-Eijnden, and Albergo]{potaptchik2026discreteflowmaps}
Peter Potaptchik, Jason Yim, Adhi Saravanan, Peter Holderrieth, Eric Vanden-Eijnden, and Michael~S. Albergo.
\newblock Discrete flow maps, 2026.
\newblock URL \url{https://arxiv.org/abs/2604.09784}.

\bibitem[Pynadath et~al.(2025)Pynadath, Shi, and Zhang]{pynadath2025candi}
Patrick Pynadath, Jiaxin Shi, and Ruqi Zhang.
\newblock Candi: Hybrid discrete-continuous diffusion models.
\newblock \emph{arXiv preprint arXiv:2510.22510}, 2025.

\bibitem[Roos et~al.(2026)Roos, Davis, Eijkelboom, Bronstein, Welling, İsmail~İlkan Ceylan, Ambrogioni, and van~de Meent]{roos2026categoricalflowmaps}
Daan Roos, Oscar Davis, Floor Eijkelboom, Michael Bronstein, Max Welling, İsmail~İlkan Ceylan, Luca Ambrogioni, and Jan-Willem van~de Meent.
\newblock Categorical flow maps, 2026.
\newblock URL \url{https://arxiv.org/abs/2602.12233}.

\bibitem[Sahoo et~al.(2024)Sahoo, Arriola, Schiff, Gokaslan, Marroquin, Chiu, Rush, and Kuleshov]{sahoo2024simple}
Subham Sahoo, Marianne Arriola, Yair Schiff, Aaron Gokaslan, Edgar Marroquin, Justin Chiu, Alexander Rush, and Volodymyr Kuleshov.
\newblock Simple and effective masked diffusion language models.
\newblock \emph{Advances in Neural Information Processing Systems}, 37:\penalty0 130136--130184, 2024.

\bibitem[Shi et~al.(2024)Shi, Han, Wang, Doucet, and Titsias]{shi2024simplified}
Jiaxin Shi, Kehang Han, Zhe Wang, Arnaud Doucet, and Michalis Titsias.
\newblock Simplified and generalized masked diffusion for discrete data.
\newblock \emph{Advances in neural information processing systems}, 37:\penalty0 103131--103167, 2024.

\bibitem[Soboleva et~al.(2023)Soboleva, Al-Khateeb, Myers, Steeves, Hestness, and Dey]{cerebras2023slimpajama}
Daria Soboleva, Faisal Al-Khateeb, Robert Myers, Jacob~R Steeves, Joel Hestness, and Nolan Dey.
\newblock {SlimPajama: A 627B token cleaned and deduplicated version of RedPajama}.
\newblock \url{https://cerebras.ai/blog/slimpajama-a-627b-token-cleaned-and-deduplicated-version-of-redpajama}, 2023.
\newblock URL \url{https://huggingface.co/datasets/cerebras/SlimPajama-627B}.

\bibitem[Song et~al.(2021)Song, Sohl-Dickstein, Kingma, Kumar, Ermon, and Poole]{song2020score}
Yang Song, Jascha Sohl-Dickstein, Diederik~P Kingma, Abhishek Kumar, Stefano Ermon, and Ben Poole.
\newblock Score-based generative modeling through stochastic differential equations.
\newblock In \emph{International Conference on Learning Representations}, 2021.
\newblock URL \url{https://openreview.net/forum?id=PxTIG12RRHS}.

\bibitem[Song et~al.(2023)Song, Dhariwal, Chen, and Sutskever]{song2023consistency}
Yang Song, Prafulla Dhariwal, Mark Chen, and Ilya Sutskever.
\newblock Consistency models.
\newblock In \emph{International Conference on Machine Learning}, pages 32211--32252. PMLR, 2023.

\bibitem[Touvron et~al.(2023)Touvron, Martin, Stone, Albert, Almahairi, Babaei, Bashlykov, Batra, Bhargava, Bhosale, et~al.]{touvron2023llama}
Hugo Touvron, Louis Martin, Kevin Stone, Peter Albert, Amjad Almahairi, Yasmine Babaei, Nikolay Bashlykov, Soumya Batra, Prajjwal Bhargava, Shruti Bhosale, et~al.
\newblock Llama 2: Open foundation and fine-tuned chat models.
\newblock \emph{arXiv preprint arXiv:2307.09288}, 2023.

\bibitem[Wei et~al.(2022)Wei, Wang, Schuurmans, Bosma, Xia, Chi, Le, Zhou, et~al.]{wei2022chain}
Jason Wei, Xuezhi Wang, Dale Schuurmans, Maarten Bosma, Fei Xia, Ed~Chi, Quoc~V Le, Denny Zhou, et~al.
\newblock Chain-of-thought prompting elicits reasoning in large language models.
\newblock \emph{Advances in neural information processing systems}, 35:\penalty0 24824--24837, 2022.

\bibitem[Yu et~al.(2024)Yu, Jiang, Shi, Yu, Liu, Zhang, Kwok, Li, Weller, and Liu]{yu2024metamath}
Longhui Yu, Weisen Jiang, Han Shi, Jincheng Yu, Zhengying Liu, Yu~Zhang, James Kwok, Zhenguo Li, Adrian Weller, and Weiyang Liu.
\newblock Metamath: Bootstrap your own mathematical questions for large language models.
\newblock In \emph{International Conference on Learning Representations}, volume 2024, pages 45040--45061, 2024.

\bibitem[Zheng et~al.(2025)Zheng, Gong, Zhang, Chen, Gu, Zhou, Jaitly, and Zhang]{zheng2025continuously}
Huangjie Zheng, Shansan Gong, Ruixiang Zhang, Tianrong Chen, Jiatao Gu, Mingyuan Zhou, Navdeep Jaitly, and Yizhe Zhang.
\newblock Continuously augmented discrete diffusion model for categorical generative modeling.
\newblock \emph{arXiv preprint arXiv:2510.01329}, 2025.

\bibitem[Zheng et~al.(2024)Zheng, Chen, Mao, Liu, Zhu, and Zhang]{zheng2024masked}
Kaiwen Zheng, Yongxin Chen, Hanzi Mao, Ming-Yu Liu, Jun Zhu, and Qinsheng Zhang.
\newblock Masked diffusion models are secretly time-agnostic masked models and exploit inaccurate categorical sampling.
\newblock \emph{arXiv preprint arXiv:2409.02908}, 2024.

\bibitem[Zheng et~al.(2023)Zheng, Chiang, Sheng, Zhuang, Wu, Zhuang, Lin, Li, Li, Xing, et~al.]{zheng2023judging}
Lianmin Zheng, Wei-Lin Chiang, Ying Sheng, Siyuan Zhuang, Zhanghao Wu, Yonghao Zhuang, Zi~Lin, Zhuohan Li, Dacheng Li, Eric Xing, et~al.
\newblock Judging llm-as-a-judge with mt-bench and chatbot arena.
\newblock \emph{Advances in neural information processing systems}, 36:\penalty0 46595--46623, 2023.

\bibitem[Zhou et~al.(2026)Zhou, Yang, Hu, Wang, Zhang, Zhang, Mackey, Jaakkola, Bates, and Zhang]{zhou2026coevolutionarycontinuousdiscretediffusion}
Cai Zhou, Chenxiao Yang, Yi~Hu, Chenyu Wang, Chubin Zhang, Muhan Zhang, Lester Mackey, Tommi Jaakkola, Stephen Bates, and Dinghuai Zhang.
\newblock Coevolutionary continuous discrete diffusion: Make your diffusion language model a latent reasoner, 2026.
\newblock URL \url{https://arxiv.org/abs/2510.03206}.

\end{thebibliography}

\newpage
\appendix

\startcontents[appendix]

\begingroup
\hypersetup{linkcolor=LabPrimary}
\section*{Appendix Contents}
\printcontents[appendix]{}{1}{\setcounter{tocdepth}{2}}
\endgroup
\newpage
\section{Gamma Schedule and Fitting}
\label{app:gamma_fitting}

To sample the bridge time during training, we use an approach similar in spirit to LangFlow~\citep{chen2026langflow}. Rather than sampling $t$ directly, we sample the log noise-to-signal ratio $\gamma$ induced by the Brownian bridge in~\eqref{eq:mlfm_forward}:
\[
    \gamma
    =
    \log\left(\sigma^2\frac{1-t}{t}\right).
\]
The inverse map is
\begin{equation}
\label{eqn:inverse_nsr}    
t = \frac{\sigma^2}{e^\gamma+\sigma^2}.
\end{equation}    

We clip $\gamma$ to $[\gamma_{\min},\gamma_{\max}]=[-10,6]$ to avoid numerically extreme values near the bridge endpoints and to keep sampling within the range covered by our empirical observations.

As in \cite{chen2026langflow}, the main component of our time-sampling distribution is fitted from the empirical difficulty of denoising at different noise levels. However, we found that using only a fitted component was less stable in our setting. We therefore sample $\gamma$ from the three-component mixture
\[
    q(\gamma)
    =
    0.1\,q_{\mathrm{unif}}(\gamma)
    +
    0.2\,q_{\mathrm{norm}}(\gamma)
    +
    0.7\,q_{\mathrm{fit}}(\gamma),
    \qquad
    \gamma\in[-10,6],
\]
and then convert $\gamma$ to $t$ using \eqref{eqn:inverse_nsr}. The uniform component $q_{\mathrm{unif}}$ is supported on $[-10,6]$. The normal component $q_{\mathrm{norm}}$ is a Gaussian with location $-2.5$ and scale $2.0$, with samples clipped to $[-10,6]$; these values were chosen from empirical observations of stable training regions.

The fitted component $q_{\mathrm{fit}}$ is constructed from cross-entropy diagnostics as a function of $\gamma$. This is the relevant diagnostic because the MLFM denoiser is trained with masked-token cross-entropy, and the average CE at a fixed noise level measures the remaining uncertainty in predicting the clean token. Thus, changes in CE across $\gamma$ indicate where the model gains information along the bridge.

Concretely, we bin training examples by $\gamma$ and compute the mean masked-token CE in each bin. After smoothing, we normalize this CE curve to obtain an empirical CDF-like curve on $[-10,6]$. We then fit the generalized-logistic CDF
\[
    F_{\mathrm{glog}}(\gamma)
    =
    \sigma\left(\frac{\gamma-\mu}{b}\right)^a,
\]
where $a>0$ is a shape parameter, $\mu$ is a location parameter, $b>0$ is a scale parameter, and $\sigma(\cdot)$ denotes the logistic sigmoid. The corresponding density is
\[
    f_{\mathrm{glog}}(\gamma)
    =
    \frac{a}{b}
    \sigma\left(\frac{\gamma-\mu}{b}\right)^a
    \left(
        1-\sigma\left(\frac{\gamma-\mu}{b}\right)
    \right),
\]
and we take $q_{\mathrm{fit}}=f_{\mathrm{glog}}$. We sample from this component by inverse transform sampling, using the quantile function
\[
    F_{\mathrm{glog}}^{-1}(u)
    =
    \mu
    -
    b\log\left(u^{-1/a}-1\right),
    \qquad
    u\in(0,1).
\]

The CE--$\gamma$ summaries are updated every 200 batches using exponential smoothing with coefficient $0.999$. We accept a new generalized-logistic fit only when its fit quality satisfies $R^2\geq 0.95$; otherwise, we keep the previous fitted component. For numerical stability, the fitted scale is lower bounded by $0.05$, and the shape parameter is clipped to $[0.05,20]$. We also use low-discrepancy stratification~\citep{kingma2021variational} for both $\gamma$ samples and mask-ratio samples within each batch.

\newpage
\section{Additional Experimental Details}
\label{app:additional_exp_detials}
Here, we provide further details on the experimental setup used in Section \ref{sec:experiments}. Unless stated otherwise, the
settings below are shared between the MLFM adaptation and SFT phases.

\subsection{Model and Adapter Hyperparameters}
\label{app:model_adapter_hparams}

The base MDM model we adapt is the official 1028M-parameter SMDM checkpoint\footnote{More specifically, we use the official SMDM 1028M checkpoint,
\texttt{mdm-1028M-3300e18-rsl-0.01-bs-1024.safetensors}, from the
\url{https://huggingface.co/nieshen/SMDM} Hugging Face repository.} of \cite{nie2025scalingmdm}, which uses a Diff-LLaMA architecture and the LLaMA-2 tokenizer \citep{touvron2023llama}. The resulting
vocabulary size, hidden dimension, and number of transformer layers are 32000,
1792, and 20, respectively. 

The pretrained backbone and token embedding matrix are kept frozen.  We attach LoRA
adapters \citep{hu2022lora} to attention and projection modules, and use a separate additive LoRA
adapter for the output head.  The time conditioning is provided by
DiT-style AdaLN adapters \citep{peebles2023scalable}. Specifically, we apply AdaLN to the normalisation layers inside each transformer block and to the final normalisation layer before the output head. We do not apply AdaLN to the token embedding layer or to the output-head module.

\begin{table}[h]
\centering
\caption{Adapter hyperparameters.}
\label{tab:adapter_hparams}
\begin{tabular}{lccc}
\hline
Component & Rank / Width & $\alpha$ & Dropout \\
\hline
Backbone LoRA & 256 & 512 & 0.05 \\
Output-head LoRA & 256 & 256 & 0 \\
AdaLN time embedding & 256 & -- & -- \\
AdaLN hidden layer & 512 & -- & -- \\
\hline
\end{tabular}
\end{table}

Our total number of trainable parameters is 319M.

\subsection{Optimisation Hyperparameters}
\label{app:optimization_hparams}

Table~\ref{tab:optimization_hparams} lists the optimisation settings for
the adaptation and SFT phase of our MLFM.  We use the AdamW optimiser \citep{loshchilov2017decoupled} with \((0.9,0.95)\) for the $\beta$ parameters.  Weight decay is applied only to matrix-valued trainable weights. We exclude bias terms and all one-dimensional affine parameters from weight decay, including AdaLN scale/shift modulation parameters.
Learning rates are linearly warmed up and then decayed by a cosine schedule to
10\% of their peak value.  We maintain an adapter-only EMA with decay 0.999
and use EMA weights for validation.

\begin{table}[h]
\centering
\caption{Optimisation hyperparameters. We abbreviate learning rate to "LR" here.}
\label{tab:optimization_hparams}
\begin{tabular}{lcc}
\hline
Hyperparameter & Adaptation & SFT \\
\hline
Global batch size & 256 & 256 \\
Gradient accumulation & 2 & 2 \\
Optimiser steps & 200k & 50k \\
Warmup steps & 3k & 3k \\
LoRA LR & \(10^{-4}\) & \(5\times 10^{-5}\) \\
Output-head LoRA LR & \(10^{-4}\) & \(5\times 10^{-5}\) \\
AdaLN LR & \(10^{-4}\) & \(5\times 10^{-5}\) \\
Weight decay & 0.01 & 0.01 \\
Gradient clipping & 1.0 & 1.0 \\
EMA decay & 0.999 & 0.999 \\
\hline
\end{tabular}
\end{table}

The adaptation stage uses 200k optimiser updates with effective batch size 512 and maximum sequence length 1024, corresponding to approximately \(100\)B processed token positions. 
The SFT stage uses 50k optimiser updates with effective batch size 512. 
The final, realised budget is approximately \(15\)B processed token positions.

\subsection{Masking Details}
\label{app:masking_details}

\looseness=-1
We sample the masking probability \(s\) from a cosine schedule inspired by
MaskGIT~\citep{chang2022maskgit}. Specifically, we draw
\(u\sim\operatorname{Unif}(0,1)\) and set
\[
    s
    =
    \rho_{\min}
    +
    (\rho_{\max}-\rho_{\min})
    \cos\!\left(\frac{\pi u}{2}\right),
\]
with \(\rho_{\min}=0.05\) and \(\rho_{\max}=1.0\). Padding tokens, special
tokens, and invalid positions are excluded from masking.

\looseness=-1
During SFT, masking is restricted to answer tokens: prompt tokens are always
kept clean. With probability \(0.5\), the full answer span is masked. Otherwise,
the answer-token masking probability is sampled from the same cosine schedule.

\subsection{Dataset Details}
\label{app:data_details}
\looseness=-1
The adaptation stage uses SlimPajama \citep{cerebras2023slimpajama} tokenised with the LLaMA-2 tokenizer \citep{touvron2023llama} using sequences of length 1024.  We use the provided train and validation
splits.  During adaptation, with probability \(0.01\) all sequences in the batch are cropped to a shared random prefix length sampled uniformly from \(\{1,\ldots,1024\}\).

\looseness=-1
The SFT stage uses a mixture of general instruction, math, and code data, with the mixture weights being 0.20, 0.60 and 0.20 respectively.  The general instruction data come from first-turn ShareGPT\footnote{The dataset can be accessed from \url{https://sharegpt.com/}.} conversations.  The math data come from the
GSM8K-Aug-NL \citep{cobbe2021gsm8k,deng2023implicit}, MetaMathQA \citep{yu2024metamath}, and NuminaMath-CoT \citep{numina_math_datasets} datasets with sub-mixture weights 0.60,
0.20 and 0.20 respectively. The code data  come from the short
OpenCodeInstruct \citep{ahmad2025opencodeinstruct} split, with the prefix fraction sampled from
\([0.25,0.75]\). SFT examples are capped at 1024 tokens, with dynamic
cropping to multiples of 64.  All SFT sources use the chat-style prompt
\verb|USER:\n{prompt}\nASSISTANT:\n|, followed by the response and an EOS
token.  Math targets append a separate answer field as
\verb|### {answer}| when available; code-instruction examples include sampled
unit tests in the prompt when available. 

\looseness=-1
SFT examples are truncated to a maximum length of 1024 tokens and 512 for MT-Bench and GSM8K respectively. To reduce
padding, batches are dynamically cropped to sequence lengths that are multiples
of 64. All SFT datasets use the same Vicuna prompt template~\citep{vicuna2023},
\verb|USER:\n{prompt}\nASSISTANT:\n|, followed by the response and an EOS
token. For math examples, when a separate final-answer field is available, we
append it to the target as \verb|### {answer}|. For code-instruction examples,
when unit tests are available, we include a sampled subset of them in the
prompt.

\subsection{SFT Details}
\label{app:sft_details}
\looseness=-1
The SFT stage is initialised from the EMA-smoothed adapter weights obtained after adaptation. We reset the optimizer and learning-rate schedule, keep the pretrained backbone and token embeddings frozen, and train only the LoRA adapters and AdaLN time-conditioning parameters. 

\looseness=-1
During SFT, masking and continuous noising are applied only to response tokens; prompt tokens are always kept clean and visible. With probability 0.5, all response tokens are masked. Otherwise, the response mask ratio is sampled from the same MaskGIT cosine schedule used during adaptation.

Additionally, we also alter the \(\gamma\)-sampling distribution during SFT as we find that the approach for $\gamma$ used during adaptation does not work as well for SFT due to the shift in the learning objective.
In particular, unlike the adaptation stage, which uses a fitted generalized-logistic component, SFT uses
an active empirical gamma curve.  The SFT sampler draws
\[
    q_{\mathrm{SFT}}(\gamma)
    =
    0.1\,q_{\mathrm{unif}}(\gamma)
    +
    0.9\,q_{\mathrm{active}}(\gamma),
    \qquad
    \gamma\in[-10,6],
\]
with a normal component \(\mathcal N(-2.5,2.0^2)\) used as a fallback until the
active curve is initialized. The active component is estimated from the high-mask
SFT diagnostics with response mask ratios in \([0.95,1.0]\).  Because SFT
contains many easy EOS and padding targets, the diagnostic curve uses
response-token CE with EOS targets removed.  We form an empirical inverse CDF
over \(\gamma\), smooth it with isotonic regression, and represent it with 101
quantile knots, and
update the active inverse CDF by EMA:
\[
    Q_{k+1}(u)
    =
    (1-\eta)Q_k(u)+\eta\,\widehat Q_k(u),
    \qquad
    \eta=0.05 .
\]
An update is applied only after at least 8 populated gamma bins and 4096
diagnostic examples are available. \cref{alg:mlfm_sft_step} summarizes one SFT learning step.

\newpage
\section{Algorithm Blocks}
\label{app:alg_blocks}

Algorithms \ref{alg:mlfm_pretraining_step} and \ref{alg:mlfm_sft_step} summarise the corresponding MLFM learning steps for adaptation and supervised fine-tuning.

\begin{algorithm}[!h]
\caption{One MLFM Adaptation Step}
\label{alg:mlfm_pretraining_step}
\begin{algorithmic}[1]
\REQUIRE Clean-text distribution \(\mathcal D\), model \(p_\theta\), mask
distribution \(\pi_s\), corruption-level distribution \(\pi_t\), optimizer.
\STATE Draw a minibatch \(X\sim\mathcal D\).
\STATE Sample a mask pattern \(\mathcal M_s\) using \(s\sim\pi_s\), and set
\(\mathcal U_s=[L]\setminus\mathcal M_s\).
\STATE Sample \(\gamma\sim\pi_t\) and then set $t$ using~\eqref{eqn:inverse_nsr}.
\STATE Construct \(z_{s,t}\): set \(z_{s,t}^\ell=E_{X^\ell}\) for
\(\ell\in\mathcal U_s\), and sample \(z_{s,t}^\ell\mid X^\ell\) from
\eqref{eq:mlfm_forward} for \(\ell\in\mathcal M_s\).
\STATE Compute
\[
    \widehat{\mathcal L}_{\mathrm{MLFM}}
    =
    \frac{1}{|\mathcal M_s|}
    \sum_{\ell\in\mathcal M_s}
    -\log p_\theta^\ell(X^\ell\mid z_{s,t},t).
\]
\STATE Update \(\theta\) with one optimizer step on
\(\widehat{\mathcal L}_{\mathrm{MLFM}}\).
\STATE Update the distribution \(\pi_t\) using the batch diagnostics at
\(\gamma\).
\STATE \textbf{return} updated \(\theta\) and \(\pi_t\).
\end{algorithmic}
\end{algorithm}

\begin{algorithm}[!h]
\caption{One MLFM Supervised Fine-Tuning Step}
\label{alg:mlfm_sft_step}
\begin{algorithmic}[1]
\REQUIRE Prompt--answer dataset \(\mathcal D_{\mathrm{FT}}\), model
\(p_\theta\), response-mask distribution \(\pi_s^{\mathrm{FT}}\),
corruption-level distribution \(\pi_t\), optimizer.
\STATE Draw a minibatch \((p,a)\sim\mathcal D_{\mathrm{FT}}\) and tokenize it
as \(X=X(p,a)\).
\STATE Let \(\mathcal P\) be prompt positions and \(\mathcal A\) be response
positions.
\STATE Sample a nonempty response mask
\(\mathcal M_s\sim\pi_s^{\mathrm{FT}}(\cdot\mid\mathcal A)\), and set
\(\mathcal U_s=[L]\setminus\mathcal M_s\).
\STATE Sample \(\gamma\sim\pi_t\) and set \(t\) using~\eqref{eqn:inverse_nsr}.
\STATE Construct \(z_{s,t}\): keep all positions in \(\mathcal U_s\) clean,
including every prompt position in \(\mathcal P\), and sample
\(z_{s,t}^\ell\mid X^\ell\) from~\eqref{eq:mlfm_forward} for
\(\ell\in\mathcal M_s\).
\STATE Compute the response-token objective
\[
    \widehat{\mathcal L}_{\mathrm{FT}}
    =
    \frac{1}{|\mathcal M_s|}
    \sum_{\ell\in\mathcal M_s}
    -\log p_\theta^\ell(X^\ell\mid z_{s,t},t).
\]
\STATE Update \(\theta\) with one optimizer step on
\(\widehat{\mathcal L}_{\mathrm{FT}}\).
\STATE Update the distribution \(\pi_t\) using the response-token diagnostics
at \(\gamma\).
\STATE \textbf{return} updated \(\theta\) and \(\pi_t\).
\end{algorithmic}
\end{algorithm}

\newpage
\section{Additional Results}
\label{app:additional_results}

\subsection{Samples}
\label{app:samplse_llm}

Figures \ref{fig:gen-example-mt-bench} and \ref{fig:gen-example-gsm8k} show two qualitative examples of MLFM performing conditional generation on MT-Bench and GSM8K respectively. 
These examples complement our quantitative results by demonstrating that MLFM can successfully handle mathematical reasoning and instruction-following tasks, which have largely remained outside the scope of prior FLM evaluations due to their inability to handle conditional generation.

\begin{figure}[!h]
\centering
\begin{genbox}[width=\linewidth, fontupper=\small]{MT-Bench}{}
\genprompt{Compose an engaging travel blog post about a recent trip to Hawaii, highlighting cultural experiences and must-see attractions.}
\genanswer{
As I returned to the tropical state of California, I was struck by the incred diversity and beauty of the islands. I arrived in the island and met with the people of my country, and I was amazed by the beauty of the blue beaches, the lush green plants, and the crystal blue waters.

One of the most famous attractions of Hawaii is the unique natural beauty of the island. From the red mountains to the tropical forests, the island is a master of natural beauty. The Kilauea's Palace, which is the largest in the United States, serves as a museum of science and technology.

As the sunseted over the island, I was curious about the rich culture of the Hawaii. The Duke Palace is the oldest and the largest museum of the islands. It is the largest collection of historical and cultural art in the world, and it is a stunning destination for tourists.

As the sunseted in the mountains, I had a fresh perspective on the history of Hawaii, a country with a rich history in the Pacific. One of the most well-known attractions in the world is the Hualaiian Cathedrals, located in the footsteps of the island's Mau National Park. The park is located in the tropical mountains, with a beautiful lana and views on the ocean.

In addition, there was a diverse culture in the island, with different customs and traditions. For example, the fire dance is a popular dance, and the Hawaii are also interested in the cone dance, which is a traditional dance that still maintains in the culture of the islands.

Another place to visit is the Hawaiolian National Zoo. It is the largest animal park in the world, where a visitor can see a variety of animals from around the world.

Overall, my trip to Hawaii was a unique experience that cannot be beat, with its unique attractions, beautiful natural beauty, and diverse cultural experiences. If you are interested in the culture of the islands, I would recommend the Hawaii Museum and the Hawaiolian National Museum, and if you want to learn more, this assistant is very helpful.
}
\end{genbox}
\caption{Qualitative example of MLFM performing conditional generation on MT-Bench.}
\label{fig:gen-example-mt-bench}
\end{figure}

\newpage
\begin{figure}[t]
\centering
\begin{genbox}[width=\linewidth, fontupper=\small]{GSM8K}{}
\genprompt{In a dance class of 20 students, 20\% enrolled in contemporary dance, 25\% of the remaining enrolled in jazz dance, and the rest enrolled in hip-hop dance. What percentage of the entire students enrolled in hip-hop dance?}
\genanswer{
20\% of 20 students is 20*0.2 = 4.
The remaining number after enrolling in contemporary dance is 20-4 = 16 students.
25\% of 16 students enrolled in jazz dance, which is .25*16 = 4 students.
So, the number of students who are enrolled in hip-hop dance is 16-4 = 12.
Therefore, the percentage of students enrolled in hip-hop dance is 12/20*100 = 60
}
\end{genbox}
\caption{Qualitative example of MLFM performing conditional generation on GSM8K.}
\label{fig:gen-example-gsm8k}
\end{figure}

\newpage
\cleardoublepage
\section{Proofs}
\label{sec:proofs}
\subsection{Proof of Proposition \ref{prop:mask_endpoint_equivalence}}\label{proof:mask_endpoint_equivalence}

\begin{proof}
At \(t=0\), every corrupted coordinate is equal to the deterministic mask
embedding \(m\), while every uncorrupted coordinate is equal to its clean
embedding \(E_{X^\ell}\).  Therefore \(z_{s,0}\) contains exactly the same
information as the partially masked sequence \(X_s\), up to the deterministic
embedding map.  Conditioning on either representation induces the same
posterior over masked token identities.
\end{proof}

\subsection{Proof of Proposition \ref{prop:promotion_error}}\label{proof:promotion_error}

Couple \(X\sim p\) with the promoted sampler. Let
\[
E
=
\left\{
    \exists i,\ \exists \ell\in P_i:
    \widehat X_i^\ell \neq X^\ell
\right\}
\]
be the event that some promoted token is wrong. Since \(P_i\) is
\(\mathcal F_i\)-measurable,
\[
\begin{aligned}
\mathbb P(E)
&\le
\sum_i \sum_{\ell=1}^L
\mathbb P\!\left(
    \ell\in P_i,\,
    \widehat X_i^\ell \neq X^\ell
\right)  \\
&=
\sum_i \sum_{\ell=1}^L
\mathbb E\!\left[
    \mathbf 1_{\{\ell\in P_i\}}
    \mathbb P\!\left(
        \widehat X_i^\ell \neq X^\ell
        \mid \mathcal F_i
    \right)
\right]  \\
&\le
\varepsilon\,
\mathbb E\!\left[\sum_i |P_i|\right]
\le
\varepsilon L ,
\end{aligned}
\]
because each coordinate is promoted at most once.

On \(E^c\), every promoted value agrees with the corresponding coordinate
of \(X\). By exact conditional dynamics and exact terminal decoding, the
remaining randomness can be coupled so that the promoted sampler finishes
with output \(\widetilde X=X\). Therefore
\[
    \operatorname{TV}(p,\widetilde p)
    \le
    \mathbb P(\widetilde X\neq X)
    \le
    \mathbb P(E)
    \le
    \varepsilon L .
\]

\end{document}